\documentclass[journal]{IEEEtran}  

\IEEEoverridecommandlockouts                              

\usepackage[utf8]{inputenc}
\let\proof\relax

\usepackage{amsmath}
\usepackage{enumerate}
\usepackage{amsfonts}
\usepackage{amssymb}
\usepackage[normalem]{ulem}
\usepackage{color}
\usepackage{xcolor}
\usepackage{comment}
\usepackage{mathtools}
\usepackage{url}
\usepackage{cite}
\usepackage{algorithm}
\usepackage{algpseudocode}
\usepackage{xfrac}
\usepackage{subcaption}
\usepackage{bm}

\DeclareMathOperator{\EX}{\mathbb{E}}
\usepackage{amsthm}

\newtheorem{theorem}{Theorem}[section]

\newtheorem{define}[theorem]{Definition}
\newtheorem{example}{Example}
\newtheorem{assumption}{Assumption}

\title{\LARGE \bf
Ensuring Reliable Robot Task Performance through Probabilistic Rare-Event Verification and Synthesis}

\author{Guy Scher$^{1}$, Sadra Sadraddini$^{2}$, Ariel Yadin$^{3}$ and Hadas Kress-Gazit$^{1}$
\thanks{$^{1}$ Sibley School of Mechanical and Aerospace Engineering, Cornell University,
        Ithaca, NY 14850, USA
        {\tt\small \{gs679,hadaskg\}@cornell.edu}}%
\thanks{$^{2}$ Dexai Robotics,
        Boston, MA, USA
        {\tt\small sadra@dexai.com}}
\thanks{$^{3}$ Department of Mathematics, Ben Gurion University of the Negev {\tt\small yadina@bgu.ac.il}}        
}

\begin{document}

\maketitle
\thispagestyle{empty}
\pagestyle{empty}


\begin{abstract}

Providing guarantees on the safe operation of robots against edge cases is challenging as testing methods such as traditional Monte-Carlo require too many samples to provide reasonable statistics. Built upon recent advancements in rare-event sampling, we present a model-based method to verify if a robotic system satisfies a Signal Temporal Logic (STL) specification in the face of environment variations and sensor/actuator noises. Our method is efficient and applicable to both linear and nonlinear and even black-box systems with arbitrary, but known, uncertainty distributions. For linear systems with Gaussian uncertainties, we exploit a feature to find optimal parameters that minimize the probability of failure. We demonstrate illustrative examples on applying our approach to real-world autonomous robotic systems.
\end{abstract}


\maketitle
\section{Introduction}
\label{sec:intro}

\IEEEPARstart{I}{n} recent years, model-based and data-driven algorithms enabled robots, and autonomous cyber-physical systems in general, to operate in the wild in unstructured environments for longer periods of time without human intervention \cite{Mero2022survey,Mascaro2021automating,feng2022robot}. Unfortunately, the performance of these systems is greatly affected by the data we train the systems on - the training distribution. For example, one cannot necessarily count on an autonomous vehicle (AV) to work as expected having to operate in a different context than it was trained on (the \emph{frame problem} \cite{Shanahan2016frame}) and it is impossible to gather data from every possible road and in any possible condition and state of the environment (pedestrians, other vehicles, visibility conditions, sensing errors, etc). This is also often called the ``open domain" problem, where the variability of the world is infinite. Yet, as the designers and manufacturers of these systems, we would like to evaluate the performance even in unseen conditions and certify that the robots will do their tasks safely.

To ensure safety in critical applications, such as autonomous driving, systems go through extensive testing phases before being deployed. Apart from physical testing, a model-based approach is indispensable, to broaden the coverage of tested scenarios. Safety requirements are often expressed using formal specification languages to reduce ambiguity; the system and environment are modeled along with the uncertainty distributions and are checked against the specification. This verification process typically attempts to find failure modes that are either examples of behaviors that violate the specification or the probability of failure. If it cannot find one, or if the probability is beneath a requisite level, the system is considered safe. 

While control theory (especially robust control), formal methods, and optimization can provide guarantees and formal certificates for these safety-critical systems, they are usually applied to systems with known Gaussian or Uniform noise distributions. Computing failure probabilities becomes intractable to compute exactly, especially when considering real-world applications such as AV, for reasons such as: non-linear dynamics; complex uncertainty distributions; dynamic and uncertain environments that interact with the system under test; long horizons; complex software components, such as machine learning algorithms; and the need to obey some complex high-level specification. 

A commonly used approach to solve these problems is using Monte Carlo (MC) simulations. The idea is to randomly sample the uncertainties in the system under test, roll out the simulations, and analyze the distribution of outcome trajectories. However, as the system is performing better and better, the probability of failing decreases, and it is harder to estimate the probability of failing with a tight confidence bound. This is called the problem of rare events and MC is inefficient and may require an exponential number of samples to correctly validate (``curse of rarity" \cite{bucklew2004introduction,liu2022curse}). Failure modes are also hard to find because we search for a sequence of decisions over long trajectories, so the search space becomes a limiting factor. And finally, we are looking for all of the possible ways to violate the specification which also increases combinatorially with time and propositions.

This paper has two main goals. The first is to efficiently and accurately estimate the probability $p^\varphi_{f}$, that the robot would fail to satisfy a specification $\varphi$. The second goal is given a system, synthesize new controllers or nominal reference trajectories such that the probability of failure will decrease. Central to our approach is the ability to sample \textit{new} specification-failing trajectories or noise combinations, that could potentially be further investigated to extract other insights. 

The problem of verifying (black-box) systems has been studied extensively \cite{corso2021survey} and is usually divided into: falsification - finding the uncertainties or disturbances that would cause the system to fail; finding the most likely failures; and, finding the probability to fail. We focus on the last part (albeit the framework can extract the first two as well) because it captures better the likelihood of the failures which can be used for certifying the system \cite{majumdar2020should}. When considering synthesis, minimizing the probability of failure is, in a way, optimizing performance over all possible scenarios, ensuring comprehensive performance and safety. 

\noindent \textbf{Contributions}: {we build upon and aggregate our previous work in \cite{scher2022elliptical,scher2023nonlin,scher2022synth} to present a Markov Chain Monte Carlo (MCMC) technique to sample and compute the probability of failing a Signal Temporal Logic \cite{donze2010robust} specification which is most useful when dealing with rare events. The technique is derived for linear systems with Gaussian noises and then extends and relaxes the limitations to non-linear and black-box systems (where only some outputs can be observed yet the state of the robot is hidden) with known arbitrary noise distributions. The technique is hyperparameter-free, meaning no special calibrations, training, or hand tuning is necessary for it to work for different problems. The technique allows us to sample new $\varphi$-failing disturbances to gain insight into system performance and to synthesize new controls that drive the probability of failure down. In this work, we extend the previous work by (1) providing another algorithm for the black-box sampler, which we call the Lipschitz-based sampler, that aims to provide a faster sampler implementation while maintaining the Markov chain properties. (2) We bring new theoretical results regarding the bounds of the multiplicative error on the computed probability for linear systems. These results may be used in conjunction with the computed probability, to assess the maximum risk possible.} 

\subsection{Related work}
\label{sec:relwork}

\subsubsection{Verification / Validation}
one approach to validation is with \textbf{formal model checking} \cite{agha2018survey,clarke2018handbook}. The idea is to use the mathematical model of the system and to prove whether specification-violating examples exist or not. These methods require full knowledge of the system, which might not be a valid assumption for complicated robots. Also, these methods do not scale well when dealing with the problems mentioned for large and complicated robotic systems \cite{alur2015principles}.

At times, we cannot assume anything about the system either because we do not have access to the model or it is too complicated to reason about and this is typically referred to as black-box models \cite{peled1999black}. A widely-used approach is to rely on \textbf{Monte Carlo} simulations \cite{rubinstein2016simulation,bucklew2004introduction} to find events that violate the specification \cite{abbas2013probabilistic}. These methods rely on sampling executions of the system in its environment, which may need an extremely large number of samples when dealing with stable and mature robotic systems. 

There are four main categories of algorithms in the literature that are used to find possible failing examples in a more directed manner. The first, \textbf{optimization} based algorithms, attempt to find the minimal cost trajectory over the distribution of possible uncertainties solely based on the output signal that the simulator is providing. The cost may involve some likelihood model of the uncertainties and or the robustness of the system. While black-box systems provide no (computationally cheap) access to gradients, there exist methods to optimize without gradients \cite{kochenderfer2019algorithms,abeysirigoonawardena2019generating,mathesen2019falsification,deshmukh2017testing,corso2020scalable,aerts2018temporal}. Two major drawbacks to this method: (a) having to search over the distribution of uncertainties, which is large, thus the methods scale poorly. (b) these optimization problems are non-linear, constrained and non-convex, hence one can be stuck in local minima and solutions are not necessarily guaranteed. Some techniques have been shown to overcome the last drawback, such as works applying ant colony optimization \cite{annapureddy2010ant}, simulated annealing \cite{abbas2013probabilistic, aerts2018temporal}, genetic algorithms \cite{zou2014safety} and even Bayesian optimization \cite{deshmukh2017testing}, but at higher computational burdens.

The second category used to find failing trajectories is using ideas taken from the \textbf{path planning} literature \cite{frances2017purely,branicky2006sampling,koschi2019computationally,dreossi2015efficient,zutshi2014multiple,tuncali2019rapidly}. Here, the algorithms search for a path that takes the system from its initial state to the set of states that violate the specification through a series of noise actions applied to the robot or simulation. These algorithms usually also scale poorly as the trajectory horizon increases. The third category, \textbf{reinforcement learning} (RL), \cite{koren2019efficient,delmas2019evaluation,lee2020adaptive,wicker2018feature} learns a policy that would maximize future rewards by taking actions (noises) that would change the state to a state that would cause the violation of the specification. These methods require access to the internal state of the simulation (i.e. not truly black-box) and RL may also pose difficulties in efficient sampling and training. The techniques mentioned thus far focus on the falsification problem, i.e. finding a failure event or finding the most likely failure event.

As previously advocated, to certify that a robot is safe, we should estimate the probability of failure because building a real robotic system that never fails is improbable. However, estimating this probability with uniform random sampling such as MC may be intractable when dealing with rare events. Hence, in the fourth category, the sampling efficiency is improved by using a class of techniques called \textbf{importance sampling} \cite{dreossi2019verifai, sinha2020neural,o2018scalable}. The idea is to skew the sampled distribution in such a way that would increase the likelihood to find failures while reducing the variance of the estimated probability that will ultimately require fewer samples. Fewer samples are critical because every evaluation of the simulation is considered costly in terms of the time to run. Then, they re-weigh the results to get an unbiased estimate of the true probability. However, it is not always trivial to choose a good importance distribution. And in fact, a badly chosen distribution can make the variance even worse \cite{glasserman1997counterexamples}.

It is worth mentioning three toolboxes for falsifying systems with respect to (w.r.t) temporal logic specifications that have become common, at least in the academic world. The toolbox S-TaLiRo\footnote{\url{https://sites.google.com/a/asu.edu/s-taliro/}} \cite{hoxha2014towards}, Breach\footnote{\url{https://github.com/decyphir/breach}} \cite{donze2010breach} and VerifAI\footnote{\url{https://github.com/BerkeleyLearnVerify/VerifAI}} \cite{verifai-cav19} take a black-box simulation or function, the parameters to falsify and a temporal logic specification and converts it to a cost function, based on the robustness metric, for the optimization program. Then, using various optimization and sampling techniques, attempt to find falsifying parameters or counterexamples. These tools do not provide an estimate to the probability of failure. Trustworthy AI\footnote{\url{https://trustworthy.ai}} provides a proprietary API (application programming interface)  through which a system may be explored to find the counterexamples and obtain an estimate of the probability to fail some safety function \cite{sinha2020neural,o2018scalable}. It is achieved by adaptive importance sampling and adaptive multilevel splitting \cite{cerou2019adaptive} to efficiently search the parameter space, but is not available to the general public. To the best of our knowledge, this is the only tool available, academic or commercial, that provides an estimate of the probability of failure (other than running MC, of course).

\subsubsection{Synthesis}
it has been a long-term goal of the robotic community to automatically synthesize controllers that will meet the robot's specifications \cite{kress2018synthesis, fainekos2009temporal}. The gains are clear - stable and provably-correct controls are automatically generated whenever an inexperienced individual changes an easier-to-reason-about, high-level, logical specification. In the context of this work, once we estimated the probability of failure, we may not be content with the results and wish to improve them. The designer may have several degrees of freedom to change in the system such as controls, paths, physical parameters, etc. In this section, we present the literature on synthesizing controls for systems with uncertainties that meet specifications, particularly for Signal Temporal Logic as we will enumerate its advantages in Sec.~\ref{sec:prelim_stl}.

The first approach that has shown tremendous success in the literature is \textbf{optimization}-based. The mathematical specification is usually transferred to a Model Predictive Control (MPC) formulation, and more specifically to a Mixed Integer Program (MIP), \cite{raman2014model, raman2015reactive,farahani2015robust,farahani2018shrinking,sadraddini2015robust,safaoui2020control, buyukkocak2021control}. The specification's predicates are encoded as binary variables that must hold. The mathematical program is, therefore, usually, hard to solve especially as the problem scales up, i.e. for larger state dimensions, time horizons, and more complicated specifications, where the number of binary variables also increases exponentially with the specification. In \cite{belta2019formal}, the authors showed how to generate robust trajectories without introducing more binary variables. We note that maximizing a reward function (e.g. the robustness), is essentially dealing with the nominal scenario, rather than minimizing the probability of failure, providing a holistic guarantee that the system will work as much as possible regardless of the disturbances or unknowns. \textbf{Control Barrier Functions} (CBF) \cite{lindemann2017robust,lindemann2018control,gundana2021event,lindemann2019decentralized} are another type of optimization techniques that works in continuous space and time, however, there is no clear notion and use of the robustness of the system, meaning even marginally-safe trajectories are acceptable. The CBF approaches also usually assume perfect state knowledge, which becomes problematic in real-world scenarios with noisy measurements, disturbances, or other parametric uncertainties.

Another line of work can be categorized as \textbf{automata}-based algorithms  \cite{kloetzer2008fully, wolff2013automaton}. However, they require the system to be abstracted so they lose the notion of the continuous space and time. 
\section{Problem setup}
\label{sec:prob_setup}

\subsection{Definitions}
\label{sec:prob_setup_def}

\noindent We consider a robotic system $\mathcal{S}$ with the following transition model:
\begin{align}
    \label{eqn:cps_system}
    x_{t+1} = f(x_{t}, u_{t}, w_{t+1}),
    ~~{y_t} = g(x_t, u_t, w_{t+1})
\end{align}
where $x_t \in \mathbb{R}^n$ is the state, $u_t \in \mathbb{R}^m$ is the control input, $y_t \in \mathbb{R}^q$ is the output signal for all $t$, and ${\bm w}=[w_0,w_1,\ldots]' \in \mathbb{R}^l$ is the uncertainty parameter input. ${\bm w}$ can affect both the process $f$ as disturbance and the output $g$ as measurement noise. With a slight abuse of notation, we denote the initial condition $x_0$ as a function of $w_{0}$ so we can assign uncertainty to it as well. 

We evaluate the system $\mathcal{S}$ w.r.t the safety specification $\varphi$ while operating in some environment $\mathrm{E}$. The specification is evaluated over the signal ${\bm y}=[y_0,y_1,\ldots]$ that is outputted from the system, the state of $\mathcal{S}$ is not necessarily exposed. The uncertainties ${\bm w}$ are chosen arbitrarily by the environment from some known probability density $\mathcal{P}({\bm w})$. Note that the density distribution cannot be state-dependent, i.e. $\mathcal{P}({\bm w}|x)$ is not allowed. 

Given known external controls $u_{0:N-1}$, a run of the system for $N$ discrete time steps is only a function of the uncertainties, i.e.:
\begin{equation}
\label{eqn:cps_run_func}
    \bm y=y_{0:N-1} = r({\bm w}), 
\end{equation} 
and given a specific $\bm w$, two runs would be identical. We call $r : \mathbb{R}^{l} \rightarrow \mathbb{R}^{qN}$ the \emph{run} function. 

\subsection{Specification requirements}
\label{sec:prob_setup_safety}
\noindent We use Signal Temporal Logic (Sec.~\ref{sec:prelim_stl}) to construct the safety specification $\varphi$. We say that if the output $\bm y$ from system $\mathcal{S}$ is satisfying $\varphi$ then ${\bm y}\in\varphi$, and ${\bm y}\notin\varphi$ if it violates it. The specification time bounds are finite, i.e. $N<\infty$. 

\subsection{Objectives}
\label{sec:prob_setup_goal}
\noindent 1. Evaluate the probability that $\mathcal{S}$ will fail the specification $\varphi$:
\begin{equation}
    p_f^\varphi=\mathbb{E}\left[ {\bm 1}\left[r({\bm w})\notin \varphi\right] \right] 
\end{equation} 
Where ${\bm 1}$ is the indicator function which is equal to $1$ if its argument is true, and $0$ if it is false.

\noindent 2. Find parameter $\gamma^*$ that (locally) minimizes the probability that $\mathcal{S}$ violates $\varphi$:
\begin{equation}
    \gamma^*=\underset{\gamma}{\mathrm{argmin}} ~p^\varphi_f
\end{equation}
Where $\gamma$ can be any parameter in the system's components, such as a controller, reference trajectory, some physical parameter, etc.

\noindent Note: we normally discuss finding failing examples in this paper because of the robotic context, but it is completely applicable to finding the probability to be successful, $p_s^\varphi$. We can always negate the specification $\varphi$ to meet the needs.
\section{Preliminaries}
\label{sec:prelim}

\subsection{Signal Temporal Logic}
\label{sec:prelim_stl}

\noindent In our context, validation of a system while in its environment is achieved w.r.t some safety property or specification and that is the only criteria upon which it is measured. There exist formal languages that allow the expression of clear and concise safety requirements as opposed to natural languages used by humans which are expressive but may produce ambiguity. Some formal languages even provide a numerical evaluation of how ``robustly" the safety requirement is achieved, referred to as quantitative semantics. 

Due to the nature of our problem where we deal with trajectories - i.e. states or outputs of the robot executions over time, a reasonable and commonly used type of formal language is called temporal logic. Temporal logic describes the properties of signals (trajectories) over time and allows us to reason about both space and time (temporal). There exist several temporal logic languages including Linear Temporal Logic (LTL) \cite{pnueli1977temporal}, Metric Temporal Logic (MTL) \cite{koymans1990specifying}, Signal Temporal Logic (STL) \cite{maler2004monitoring,donze2010robust} and more.

In this work, we utilize STL to create specifications. An STL specification is comprised of atomic propositions, called predicates, $\mu$ that map an array that is defined by $n$ variables, $s_t\in \mathbb{R}^n, t\in \mathbb{N}_+$ to a real value, $\mu : \mathbb{R}^n \rightarrow \mathbb{R}$. A predicate $\mu$ is considered true if $\mu(s_t)\geq 0$, and false otherwise. We consider discrete-time real-valued signals ${\bf s}=s_0,s_1,\ldots$. The STL grammar is given by \cite{donze2010robust}:
\begin{align}
    \varphi := \mu~|~\neg\varphi~|~ \varphi_1 \wedge \varphi_2~ |~ \varphi_1\, \mathcal{U}_\mathcal{I}\, \varphi_2,
\end{align}
where $\varphi, \varphi_1$ and $\varphi_2$ are STL formulae and $\mathcal{I}=[t_1,t_2]$ is an interval over which the property is considered, where $0\leq t_1<t_2<\infty$. The Boolean operator $\wedge$ is conjunction (\textit{and}) and $\neg$ is negation. $\varphi_1\,\mathcal{U}\,\varphi_2$ means $\varphi_1$ holds true \textit{until} $\varphi_2$ becomes true in the future signal. Other Boolean and temporal operators can be constructed from these, such as $\vee$ - disjunction (\textit{or}), \textit{always} $\Box\varphi$ ($\varphi$ holds on the entire successive signal) and \textit{eventually} $\Diamond\varphi$ ($\varphi$ becomes true at some point in the successive signal). The satisfaction of the STL formula $\varphi$ by the signal $\bf s$ is denoted by $({\bf s}, t) \models \varphi$.

\begin{example}
Consider a point robot in $\mathbb{R}^2$ with a state $s_t=(x_t,y_t)'$ navigating in a field. The specification:
\begin{align*}
\varphi:=\Box_{[0,100]}(\neg( &\left(x-p_x<1\right) \wedge \left(x-p_x>-1\right) \wedge \\ &\left(y-p_y<1\right)\wedge \left(y-p_y>-1\right))\,) \\ \wedge 
\Diamond_{[90,100]}(&\left(x-0<0.1\right) \wedge \left(x-0>-0.1\right) \wedge \\ &\left(y-0<0.1\right)\wedge \left(y-0>-0.1\right))
\end{align*}
means that for 100 time steps, the horizontal and vertical distance away from an obstacle located at $(p_x,p_y)'$ cannot be less than 1m at the same time, and the robot must reach the origin eventually within 0.1m over the last 10 time steps.
\label{ex:stl_spec}
\end{example}

STL provides a metric of the ``distance" to satisfaction of $\varphi$, known as the robustness metric, or the STL score. A positive robustness value $\rho$, indicates $({\bf s},t)\models \varphi$ and a negative robustness indicates the violation of $\varphi$. 

\begin{define}
The STL robustness $\rho\left({\bf s},\varphi,t\right)$ is defined recursively as:
\begin{itemize}[]
    \item $\rho({\bf s},\mu,t)=\mu(s_t)$,
    \item $\rho({\bf s},\neg \varphi,t)=-\rho({\bf s},\varphi,t)$,
    \item $\rho({\bf s},\varphi_1 \wedge \varphi_2,t)=\min(\rho({\bf s},\varphi_1,t),\rho({\bf s},\varphi_2,t))$, 
        \item $\rho({\bf s},\varphi_1 \vee \varphi_2,t)=\max(\rho({\bf s},\varphi_1,t),\rho({\bf s},\varphi_2,t))$, 
    \item  $\rho({\bf s},\Diamond_{[t_1,t_2]} \varphi, t) = \underset{\tau \in t+[t_1,t_2]}{\max} \rho({\bf s},\varphi,\tau)$, 
     \item  $\rho({\bf s},\Box_{[t_1,t_2]} \varphi, t) = \underset{\tau \in t+[t_1,t_2]}{\min} \rho({\bf s},\varphi,\tau)$,
    \item $\rho({\bf s},\varphi_1 \mathcal{U}_{[t_1,t_2]} \varphi_2, t) = \\ \underset{\tau \in t+[t_1,t_2]}{\max}( \min\big(\rho({\bf s},\varphi_2,\tau),\underset{\tau' \in [t,\tau]}{\min}\rho({\bf s},\varphi_1,\tau')))$. 
\end{itemize}
\label{def:stl_score}
\end{define}
The STL robustness of a signal and a specification is defined as $\rho({\bf s},\varphi, 0)$.
Note: to correctly evaluate $\rho({\bf s}, \varphi, 0)$, the minimum length of signal $\bf s$ must be $H^\varphi$ where:
\begin{itemize}
    \item $H^{(\mu(s)\ge 0)}=1$,
    \item $H^{\neg \varphi}=H^{\varphi}$,
    \item $H^{\varphi_1 \wedge \varphi_2}=H^{\varphi_1 \vee \varphi_2}=\max(H^{\varphi_1}, H^{\varphi_2}) $,
    \item $H^{\Diamond_{[t_1,t_2]} \varphi} = H^{\Box_{[t_1,t_2]} \varphi} = t_2+   H^{\varphi} $,
        \item $H^{\varphi_1 \mathcal{U}_{[t_1,t_2]} \varphi_2} =  t_2 + \max(H^{\varphi_1}, H^{\varphi_2})$.
\end{itemize}

\begin{define}
The $\varrho$-level set of an STL formula $\varphi$ is defined as:
\begin{equation}
    \mathcal{L}(\varphi,\varrho) = \{ {\bf s} ~|~ \rho({\bf s}, \varphi, 0) \ge  \varrho \},
\end{equation}
means the set of signals (trajectories) that have a robustness value greater than $\varrho$.
\label{def:rho_level_set}
\end{define}

\subsection{Probability and Confidence bounds}
\label{sec:prelim_prob}

\noindent In this paper, we provide  bounds on the possible error produced by our method. We list these corollaries here to reference them when needed:
\begin{enumerate}
    \item[(B1)] \label{cor:markov_ineq} Markov's inequality gives us the probability that a non-negative random variable is greater than some constant value by relating probabilities to expectations. If $X$ is a non-negative random variable and $a>0$, then:
    $$
    \mathbb{P}(X\geq a) \leq \frac{\mathbb{E}[X]}{a}
    $$
    \item[(B2)] \label{cor:hoeff_ineq} Hoeffding's lemma provides an inequality that limits the moment-generating function of a bounded random variable. If $X$ is a random variable s.t. $X \in [a,b]$, then for all scalar $\epsilon$:
    $$
    \mathbb{E}[e^{\epsilon(X-\mathbb{E}[X])}] \leq e^{\epsilon^2(b-a)^2/8}
    $$
    \item[(B3)] \label{cor:bin_dist} The binomial distribution $B(N,q)$ measures the number of successes when conducting $N$ independent experiments, where the output of every experiment is successful, or true, with the probability $q$. The expectation of $X\sim B$ is $Nq$ and the variance is $Nq(1-q)$.
\end{enumerate}

\subsection{Elliptical Slice Sampling and HDR}
\label{sec:prelim_ess_hdr}

\begin{figure}
     \centering
     \begin{subfigure}[b]{0.33\columnwidth}
         \centering
         \includegraphics[width=\columnwidth]{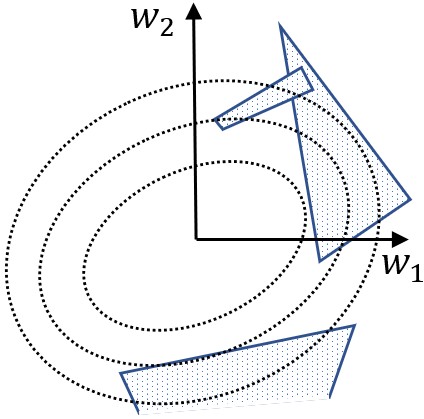}
         \caption{}
         \label{fig:lincon_gauss}
     \end{subfigure}
     \hfill
     \begin{subfigure}[b]{0.27\columnwidth}
         \centering
         \includegraphics[width=\columnwidth]{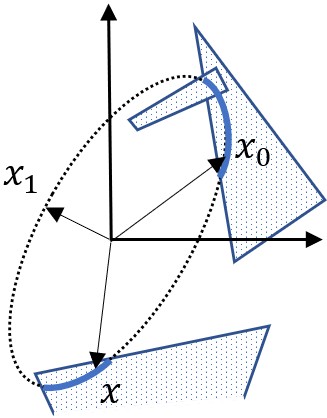}
         \caption{}
         \label{fig:ess_sample}
     \end{subfigure}
     \hfill
     \begin{subfigure}[b]{0.28\columnwidth}
         \centering
         \includegraphics[width=\columnwidth]{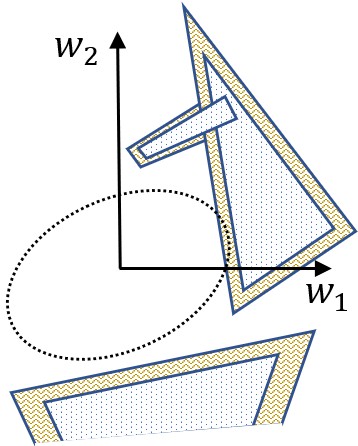}
         \caption{}
         \label{fig:hdr_phase}
     \end{subfigure}
        \caption{The pipeline overview of sampling and estimating $p_f^\varphi$ using STL-based ESS~\cite{scher2022elliptical}. (a) The uncertainty distribution $\mathcal{P}({\bm w})$ is represented with the dotted ellipses, and the constrained domains $\mathcal{L}$ that falsify the specification are shaded blue. Objective 1 is to estimate $p(x\in\mathcal{L})$ - the integral of the Gaussian over the blue polygons. (b) Description of the ESS process - given a point $x_0\in\mathcal{L}_{k-1}$, sample a new point $x\sim\mathcal{P}({\bm w})\in\mathcal{L}_{k-1}$ using the auxiliary variable $x_1$. (c) Description of the HDR phase with the inflated regions. According to Eq.~\eqref{eqn:p_prod_cond}, $p(\operatorname{blue})=p(\operatorname{blue}|\operatorname{orange\cup blue})p( \operatorname{orange\cup blue})$ (where the orange represents only the additive part).  }
        \label{fig:stl_ess_hdr}
\end{figure}

\noindent Our goal is to evaluate $p_f^\varphi$, the probability of the robot to fail the specification $\varphi$. This can be achieved by computing the integral over the set of signals (regions) where the specification is being violated, $\mathcal{L}(\varphi,0)$ as in Def.~\ref{def:rho_level_set}, over $\mathcal{P}({\bm w})$, the distribution of the uncertainties. Mathematically,  $p_f^\varphi=\mathbb{P}({\bf x}\in\mathcal{L})=p({\bf x}\in\mathcal{L})=\int_{{\bf x}\in\mathcal{L}}d\mathcal{P}$. Fig.~\ref{fig:lincon_gauss} depicts the constrained regions in blue and the distribution contours in a dotted ellipse. In general, numerical integration with quadrature methods is intractable when dealing with high dimensional spaces \cite{orive2020cubature}. There are also no analytical solutions when the bounding hyperplanes of the regions $\mathcal{L}$ are not axis-aligned with $\mathcal{P}$. If we try to estimate $p({\bf x}\in\mathcal{L})$ with naive MC sampling when the probability is very low and the dimension of ${\bf x}$ is high, we would need many samples to confidently estimate the probability, which would become very inefficient \cite{bucklew2004introduction}. It may even become infeasible and impractical when considering our problem of complicated black-box simulations that require significant time to run. 

The authors in \cite{gessner2020integrals} presented an approach to estimating the probability that a sample sampled from a Gaussian  $\mathcal{N}(\mu,\Sigma)\in\mathbb{R}^n$, is also within some \emph{single}, \emph{linearly} constrained region (truncated Gaussian), $\mathcal{L}\subseteq \mathbb{R}^n$ using an MCMC sampler. The domain $\mathcal{L}$ is defined by the intersection between all $d$ constraints, i.e. $Hx+h\ge 0$ where $H\in\mathbb{R}^{d\times n}, h\in\mathbb{R}^{d}$. The approach re-iterates between two phases: the first, is a multilevel splitting algorithm called Holmes-Diaconis-Ross (HDR) \cite{diaconis1995three} which enables evaluating even very low probabilities by creating a series of nested domains by inflating them and composing $p(x\in\mathcal{L})$ into a product of $K$ easier to compute conditional probabilities. The second phase, the MCMC, is based on Elliptical Slice Sampling (ESS) \cite{murray2010elliptical} that enables us to sample $x\sim\mathcal{N}(\mu,\Sigma)\in\mathcal{L}_k$ rejection-free from each nested domain $k=1...K$. We present the essentials of the method developed in \cite{gessner2020integrals} with this paper's notation for clarity and completeness.

\subsubsection{HDR}
\label{sec:prelim_hdr}

the algorithm's purpose is to repeatedly inflate the region $\mathcal{L}$ in such a way that $\mathcal{L}_0:=\mathbb{R}^n,~p(\mathcal{L}_0)=1$ and $\mathcal{L}_K=\mathcal{L}$ such that:
\begin{equation}
\label{eqn:p_prod_cond}
 p(\mathcal{L})=p(\mathcal{L}_0) \prod_{k=1}^K p(\mathcal{L}_k | \mathcal{L}_{k-1})   
\end{equation}
Each nesting $\mathcal{L}_k$ is inflated by a constant shift $\delta_{k-1}>\delta_{k}\geq 0$, $Hx+h+\delta_k\ge 0$, in such a way that the conditional probability $p(x\in\mathcal{L}_k | x\in\mathcal{L}_{k-1})\approx 0.5$. $N_{\text{ESS}}$ samples are sampled from each nesting $\mathcal{L}_{k-1}$ and the conditional probability can then be estimated as $p(\mathcal{L}_k | \mathcal{L}_{k-1}) = \sum_{j=1}^{N_{\text{ESS}}} {\bm 1}[x_j\in\mathcal{L}_{k}]/\sum_{j=1}^{N_{\text{ESS}}}{\bm 1}[x_j\in\mathcal{L}_{k-1}]$. See Fig.~\ref{fig:hdr_phase} for an illustration of single nesting inflation. The number of nestings $N_{\text{HDR}}$ is approximately $\lceil\log_2(1/q)\rceil$ ($q$ - the true, latent, probability of failing).

\subsubsection{ESS}
\label{sec:prelim_ess}

the ESS procedure is the means to sample from a given nesting $\mathcal{L}_{k-1}$ rejection-free. It starts with a given point $x_0\in\mathcal{L}_{k-1}$, and eventually produces another point $x\sim\mathcal{N}(\mu,\Sigma)\in\mathcal{L}_{k-1}$. The process is depicted in Fig.~\ref{fig:ess_sample}. An auxiliary variable  $x_1\sim\mathcal{N}(\mu,\Sigma)$ is sampled, forming an ellipse parameterized by a scalar $\theta$: $x(\theta)=x_0 \cos(\theta)+x_1 \sin(\theta)$. We can then sample $x(\theta),~\theta \sim U[\theta\,|\,x(\theta)\in\mathcal{L}_{k-1}]$ (uniformly from the ellipse segment that overlaps the domain $\mathcal{L}_{k-1}$). The efficiency of the approach comes from the fact that there exists a closed-form analytical solution to the intersection between the auxiliary ellipse and the \textit{linearly} constrained region $\mathcal{L}_{k-1}$, thus we can sample directly from it without doing rejecting samples. To weaken the dependency between consecutive samples $x$, a ``burn-in" procedure is introduced. Discarding $N_{\text{skip}}$ samples between the samples that are taken into account when sampling $N_{\text{ESS}}$ samples in each nesting.

In a sense, this method is allowing us to sample from the domain by removing segments on a one-dimensional ellipse contour that are not in the domain. Sampling uniformly on the 1-D known leftover domain is easy. The equivalent in higher dimensional space of our trajectories, would be to keep removing an $n$-dimensional ball about a sampled point where we a guaranteed to be outside the domain (minimal distance to the closest failing domain) and sampled from the Gaussian conditioned on the leftover regions. However, there is no known method that can sample rejection-free from such a distribution in $n$ dimensions.

\section{Verification}
\label{sec:verification_intro}

\noindent In this chapter, we solve objective 1, i.e. estimate $p_f^\varphi$. Our approach is to verify a robot w.r.t specification $\varphi$, by finding $p(r({\bm w})\notin\varphi)$. We extend the work in \cite{gessner2020integrals} to allow sampling from multiple domains simultaneously to capture all the different ways the robot can violate $\varphi$. The different regions composing $\mathcal{L}(\varphi,0)$ represent regions that for example hit obstacle 1 at $t=1$, hit obstacle 1 at $t=2$ (hitting obstacle 1 at $t=1$ and $t=2$ is the intersection between these regions), etc \ldots We start with the particular case where the system is linear(izable), the noises are Gaussian (Sec.~\ref{sec:verification_lingauss}) and the trajectory-space distribution can thus be expressed as a high-dimensional Gaussian with linearly constrained predicates. In this scenario, we can obtain ``textbook" solutions. Unfortunately, many interesting robotic systems are non-linear, the uncertainty distributions are multi-modal or non-Gaussian and the predicates are not linear (e.g. the contour of an obstacle). We then lift the constraints (Sec.~\ref{sec:verification_blackbox}) with some mild assumptions to allow the verification of nonlinear or black-box robotic systems, with arbitrary uncertainty distributions and predicates.

\subsection{Linear Systems, Gaussian Uncertainties, Linear Predicates}
\label{sec:verification_lingauss}

\subsubsection{Setup}
following the work in \cite{scher2022elliptical}, we consider discrete, time-variant linear(izable) systems of the form:
\begin{align}
\label{eqn:linsys_dyn}
    x_{t+1}&=A_t x_t + B_t u_t + w_t, \\\nonumber
    y_{t}&=C_t x_t + v_t,
    \\\nonumber
    x_0&\sim \mathcal{N}\left(\mu^x_0,\Sigma^x_0\right)
\end{align}
With time index $t$ and interval $\Delta t$, state $x_t\in\mathbb{R}^n$, controller $u_t\in\mathbb{R}^m$ and observations $y_t\in\mathbb{R}^q$. The matrices $A_t, B_t$, and $C_t$ represent the transition matrices between the states, inputs, and measurements and are assumed known. For convenience, we differentiate the uncertainties to the process noise $w_t \sim \mathcal{N}(\mu_t^w, \Sigma_t^w) \in \mathbb{R}^n$ and the measurement noise $v_t \sim  \mathcal{N}(\mu_t^v, \Sigma_t^v) \in \mathbb{R}^q$. The system may also contain an estimator:
\begin{equation}
\label{eqn:linsys_obs}
    \hat{x}_{t+1}=A_t \hat{x}_t + B_t u_t + L_t (y_t-C_t \hat{x}_t)
\end{equation}
and state- or measurement-based feedback controller with reference commands $r_t$ and controller $K_t$:
\begin{equation}
\label{eqn:linsys_ctrl1}
    u_{t}=r_t - K_t \hat{x}_t,~\text{or}~ u_{t}=r_t - K_t y_t
\end{equation}
The specification's predicates are formed with linear hyperplanes, i.e. the intersection of $d_j-$constraints, $\mu_j:=(H_jx+h_j \ge 0), H_j \in \mathbb{R}^{d_j\times n}, h_j \in \mathbb{R}^{d_j}$ for predicates $j=1\ldots N_\mu$ that form $\varphi$. 

\subsubsection{Trajectory-space}
we represent the system trajectories with a Gaussian in a higher dimensional space ${\bf x}_{\operatorname{traj}}:=[x_0',\dots,x_{N-1}']'\in\mathbb{R}^{n \cdot N}$ where the trajectory horizon $N \geq H^\varphi$. ${\bf x}_{\operatorname{traj}}$ is obtained by substituting Eq.~\eqref{eqn:linsys_dyn}-\eqref{eqn:linsys_ctrl1} for every time step. A matrix relationship can then be derived that binds the initial state $x_0$, the controls, and the process and noise measurements to the trajectory:
\begin{flalign}
\label{eqn:linsys_resp}
{\bf x}_{\operatorname{traj}} = \Phi_0 x_0 + \Phi_r {\bf r} + \Phi_v {\bf v} + \Phi_w {\bf w}
\end{flalign}
where ${\bf r}=[r_0',\dots,r_{N-1}']'\in\mathbb{R}^{m\cdot N}$, ${\bf v}=[v_0',\dots,v_{N-1}']'\in\mathbb{R}^{q\cdot N}$, and ${\bf w}=[w_0',\dots,w_{N-1}']'\in\mathbb{R}^{n\cdot N}$. $\Phi_0,\Phi_r,\Phi_v$ and $\Phi_w$ link these arrays to the trajectory. Finally, the trajectory-level Gaussian is obtained:
\begin{flalign}
\label{eqn:trajectory_pdf}
    {\bf x}_{\operatorname{traj}} \sim \mathcal{N}\big(& \Phi_{0} \mu_0^x  + \Phi_r {\bf r}+ 
    \Phi_v \operatorname{M_v} + \Phi_w \operatorname{M_w}, \\\nonumber
        &\Phi_0 \Sigma_0^x \Phi_0' + \Phi_v 
        \Sigma_v \Phi_v^\prime+ 
        \Phi_w 
        \Sigma_w \Phi_w^\prime \big)
\end{flalign}
$M_i=[\mu_0^{'i},\ldots,\mu^{'i}_{N-1}]'$, $i\in\{v,w\}$. 

\subsubsection{STL-based ESS} 
we propose a robustness-guided sampling technique to verify the robotic system w.r.t STL specifications. To find the probability of failure, the integral that we evaluate:
\begin{flalign}
\label{eqn:eq_stl_integral}
    p_f^\varphi= \underset{{\bf x}_{\operatorname{traj}} \in \mathcal{L}(\varphi,0)}{\int} \mathcal{P}({\bf x}_{\operatorname{traj}}) d{{\bf x}_{\operatorname{traj}}},
\end{flalign}
We use modified ESS and HDR algorithms to guide the sampling towards $\mathcal{L}(\varphi,0)$ and to evaluate Eq.~\eqref{eqn:p_prod_cond} using the STL robustness metric. We would like to sample rejection-free from $\mathcal{L}(\varphi,\rho_k)$, regions in the trajectory-space where the robustness is greater than $\rho_k$. However, there is a problem. Even just enumerating the union of all polyhedra that violate $\varphi$ is challenging, because it can grow exponentially in the size of the formula (see \cite{sadraddini2016feasibility} and example in the preface of Sec.~\ref{sec:verification_intro}). We avoid explicit enumeration of the polyhedra in $\mathcal{L}(\varphi,\rho_k)$ while computing the integral in Eq.~\eqref{eqn:eq_stl_integral}. To achieve this, we only need the STL robustness metric.

\begin{theorem}[\cite{scher2022elliptical}]
\label{thr:stl_ess}
Given: (1) an STL formula $\varphi$ with a set of linear predicates $\mu_j:=(H'_jx+h_j \ge 0), j=1,\ldots,N_\mu$, where $N_\mu$ is the total number of predicates. (2) an initial sample trajectory ${\bf x}_{0} \in \mathcal{L}(\varphi,\varrho)$ and second sample trajectory ${\bf x}_{1}\sim\mathcal{N}(\mu, \Sigma)$ (not necessarily in $\mathcal{L}(\varphi,\varrho)$); Construct the ellipse $\mathcal{E} = \{{\bf x}_{0} \cos \theta+  {\bf x}_{1} \sin \theta~|~\theta \in [0,2\pi]\} \in \mathbb{R}^{n\cdot N}$. Then construct the following sorted list of real numbers in $[0,2\pi]$:
\begin{equation}
\label{eqn:ess_theta_sorted}
\begin{array}{ll}
\Theta = \operatorname{sorted} \{ & \theta~|~\exists t \in \{0\ldots t_{H^\varphi-1}\},  j \in \{1\ldots N_\mu\}, \\
 \text{ s.t. } & H'_j x_t+h_j = \pm \varrho, \\
 & {\bf x} = {\bf x}_0 \cos \theta+  {\bf x}_1 \sin \theta, \\
& {\bf x}=(x_0',x_1',\ldots,x_{N-1}')'\big \}.
\end{array}
\end{equation} 
Then for any two consecutive elements $\theta_a,\theta_b \in \Theta$ (cyclic), one of the following statements is correct:
\begin{flalign}
&\forall \theta \in [\theta_a,\theta_b], \rho({\bf x}_0 \cos \theta+  {\bf x}_1 \sin \theta) \ge \varrho, \text{ or} \\
&\forall \theta \in [\theta_a,\theta_b], \rho({\bf x}_0 \cos \theta+  {\bf x}_1 \sin \theta) \le \varrho.
\end{flalign}
\end{theorem}
\proof{From \cite{scher2022elliptical},
when $\rho({\bf x})=\varrho$, at least one of the predicates in $\varphi$ must be equal to $\pm \varrho$ - this predicate becomes the maximizer or minimizer in the STL robustness function. The reason it could have $\pm$ is that negation might be in the specification. The set $\Theta$ contains all the roots for $\rho({\bf x}) - \varrho = 0$ (but can contain extra elements which are not necessarily the roots due to the $\pm$). Since $\rho$ is Lipschitz continuous, $\rho({\bf x}) - \varrho$ is sign-stable on $\mathcal{E}$ between two consecutive roots. \qed
}

\begin{figure}
     \centering
     \includegraphics[width=\columnwidth,height=4cm,keepaspectratio]{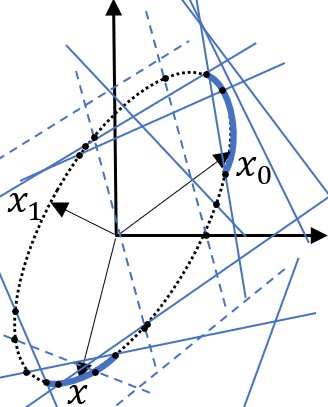}
     \caption{The hyperplanes forming $\mathcal{L}(\varphi,\varrho)$ producing the intersections with the auxiliary ellipse in the STL-based ESS. Dashed lines are predicates not in their time interval, thus an intersection point is not a root. Given $x_0$, the output is $x\in\mathcal{L}(\varphi, \varrho)$.}
     \label{fig:ess_fullstl}
\end{figure}

Theorem \ref{thr:stl_ess} enables us to efficiently extract the portions of the ellipse that fall onto $\mathcal{L}(\varphi,\varrho)$ by finding the roots of $\rho({\bf x})$ on each ellipse during the ESS process. Eq.~\eqref{eqn:ess_theta_sorted} provides all the possible roots with the worst-case computation burden of solving $2 N N_\mu$ hyperplane-ellipse intersections (which is negligible as closed-form solutions exist \cite{gessner2020integrals}). To examine whether if each segment in $\Theta\in\varphi$, we can sample a single $\theta \sim U[\theta_a,\theta_b]$ or take $\theta=\theta_a+\epsilon$ deterministically within each consecutive pair, and evaluate if ${\bf x}(\theta)\in\mathcal{L}(\varphi,\varrho)$.  Fig.~\ref{fig:ess_fullstl} illustrates the adjusted ESS procedure.
The theorem allows us to avoid the explicit representation of the $\mathcal{L}(\varphi,\varrho)$, which as previously mentioned is exponential in the length of the formula, to $\mathcal{O}(N N_\varphi)$ intersection evaluations.

To complete the second part of the algorithm, HDR, we can trivially use $\delta_k=\varrho$ with different shift values for each nesting, to inflate the regions as discussed in Sec.~\ref{sec:prelim_hdr}. Intuitively, we lower (negative) the robustness $\rho_k$ such that more trajectories satisfy the new and relaxed specification, ${\bf x} \in \mathcal{L}(\varphi, \rho_k)$. On the other hand, we must shift the hyperplanes as well with the same $\rho_k$ to find the new roots. Care must be taken in this case because a predicate may need to be inflated or deflated to be relaxed. Consider the following example, $\varphi_1:=\mu$ and $\varphi_2:=\neg \mu$. To sample ``more easily" from $\varphi_1$, we need to inflate by $-\rho_k$ so that more samples are likely to fall in it (note $\rho_k$ is negative), whereas in $\varphi_2$ we would need to deflate $\mu$ by $\rho_k$. To avoid complicated analysis of the predicates in the specification, we simply check the intersections with both $\pm\rho_k$, thus adding the spurious elements in $\Theta$ as discussed in Theorem~\ref{thr:stl_ess} (negligible overhead). Of course, an analysis of the predicates or the decision to accept only positive/negation normal form specifications, can reduce this overhead at the expense of other computations or a more restrictive specification language.

\subsection{Black-box systems, Arbitrary Uncertainties and Predicates}
\label{sec:verification_blackbox}

\noindent We now extend the approach to solve objective 1 for general black-box systems, Eq.~\eqref{eqn:cps_system}. To solve this scenario, we use similar techniques as in Sec.~\ref{sec:verification_lingauss}, with a few modifications that are necessary due to the following pains. The fact that the system is non-linear means we cannot obtain the distribution of the trajectories, $\mathcal{P}({\bf x}_{traj})$, analytically. Even if the uncertainty ${\bm w}$ is Gaussian, its propagation through non-linear dynamics does not preserve the Gaussian attributes. Approximation of $\mathcal{P}({\bf x}_{traj})$ as a Gaussian is expensive to obtain - we will need to run many simulations to capture the distribution, so at that point, we might as well estimate $p_f^\varphi$ with MC. Furthermore, it results in inherent estimation errors due to its inability to capture complicated, heavy-tailed, or multi-modal distributions. Moreover, since in general the initial uncertainty distribution $\mathcal{P}({\bm w}) \nsim \mathcal{N}$, the trajectory distribution is not Gaussian as well. Lastly, general non-linear boundary predicates mean that there is no analytic closed-form solution that can help find the segments on the ellipses that are on $\mathcal{L}(\varphi, \rho_k)$, aside from some special cases like polynomials, circles, and ellipses.

We first change the space in which we integrate to the \emph{uncertainty} space, and solve the following integral:
\begin{flalign}
\label{eqn:nonlin_stl_integral}
    p_f^\varphi= \underset{{\bf y} \in \mathcal{L}(\varphi,0)}{\int} \mathcal{P}({\bm w}) d{{\bm w}},
\end{flalign}
assuming for now that ${\bm w}\sim\mathcal{N}(\mu,\Sigma)$, a limitation that we will lift in Sec.~\ref{sec:ver_bb_nf}, and we solve this integral with STL-based ESS. The implications of this change are that the regions in ${\bm w}$ where the output of the simulation is violating the specification ${\bf y}\in\mathcal{L}(\varphi,0)$ are non-linear, non-convex, and perhaps even sparsely distributed. The domains cannot be trivially extracted from the specification as in \cite{scher2022elliptical}, because the specification is on the states (or outputs), yet we work in the uncertainty space.

There are two assumptions on the system that must hold to solve Eq.~\eqref{eqn:nonlin_stl_integral} efficiently, \cite{scher2023nonlin}:
\begin{assumption}
\label{assume_lipshitz}
The map $\rho(y_{0:N-1},\varphi,0)$ is Lipschitz continuous. 
\end{assumption}
\begin{assumption}
\label{assume_distribution}
The uncertainty source inputs ${\bm w}$ form a probability space with a \emph{known} distribution $\mathcal{P}({\bm w}): \mathbb{R}^{l} \rightarrow \mathbb{R}_+$. 
\end{assumption}
Assumption \ref{assume_lipshitz} is held when the system in Eq.~\eqref{eqn:cps_system} and predicates appearing in the STL specification are Lipschitz continuous in the state. Assumption \ref{assume_distribution} is trivial in the sense that it is only necessary for the purpose of sampling from the uncertainty distribution. Obtaining the true distribution of the uncertainties is a research topic on its own \cite{aastrom1971system,papamakarios2021normalizing} and is out of the scope of this work.

We use the same two-stage process, HDR which remains exactly the same, and ESS which is modified.

\subsubsection{ESS with Surrogate functions}
\label{sec:ver_bb_ess_surr}
the boundaries of the domains $\mathcal{L}(\varphi, \rho_k)$ are not linear in ${\bm w}$ so there is no hope of finding the exact intersections analytically, however, we must still sample efficiently from the ellipse segments that are on $\mathcal{L}(\varphi, \rho_k)$ in ESS, Eq.~\eqref{eqn:nonlin_ess_sample}:
\begin{flalign}
\label{eqn:nonlin_ess_sample}
\theta\sim U[\theta|{\bm w}(\theta)\in \mathcal{L}_{{\bm w}}(\varphi, \rho_k)]
\end{flalign}
With a slight abuse of notation where $\mathcal{L}_{{\bm w}}(\varphi, \rho_k)$ represent the regions in ${\bm w}$, where ${\bf y}=r({\bm w})\in\mathcal{L}(\varphi, \rho_k)$.
We propose two algorithms to solve Eq.~\eqref{eqn:nonlin_ess_sample}.

\noindent(a) Bayesian optimization (BO) with Gaussian Processes (GP): we use BO with GP \cite{agnihotri2020exploring} to approximate the actual robustness function of $\rho\left(r\left({\bm w}\left(\theta\right)\right), \varphi, 0\right),~\theta\in[0,2\pi]$ which we shall denote as $\varrho^\varphi\left(\theta\right)$. BO is especially useful in cases where obtaining a sample or its gradient is a costly process, and its purpose is to attempt to find either the maximal value (exploitation) or the distribution of an unknown function (exploration) in as few iterations as possible using its Bayesian inference for GP. BO also provides ``knobs" to attenuate between exploitation to exploration, which we utilize.

\begin{figure}
    \centering
    \includegraphics[scale=0.65]{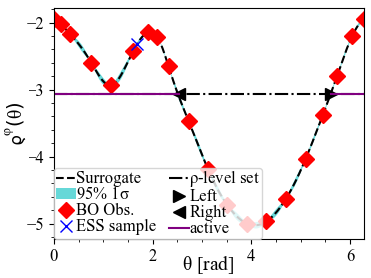}
    \caption{ 
    An example (Sec.~\ref{sec:demos_lin_ex_2dG}) of robustness values along the ellipse for linear predicates. Points sampled in the GP process are shown in red. The boundaries of the exact active segments in black triangles are computed using the hyperplane-ellipse closed-form solution. The dashed line represents the estimated $\varrho^\varphi(\theta)$, the cyan region is the uncertainty estimation, and the dash-dot line represents the shift value $\rho_{k-1}$ (for the HDR phase). The active domain in solid purple, found using the non-linear formulation \cite{scher2023nonlin}. The output, sampled from the active segment, is represented by the blue X. The estimated active domain is nearly identical to the exact domain.}
    \label{fig:lin_nonlin_comp}
\end{figure}

The algorithm is depicted here and a pseudo algorithm is listed in Algorithm \ref{alg:bayes_sample}. The input to the sampler is a given point ${\bm w}_0\in\mathcal{L}_{{\bm w}}(\varphi, \rho_{k-1})$ (the $0$ represents the given point $\in\mathbb{R}^l$, not to be confused with $w_0$ which is the initial condition uncertainty), the nesting robustness threshold $\rho_{k-1}$, the maximum number of samples $N_{\operatorname{BO}}$, the BO knobs $\kappa$ and $\xi$ that switch between exploration or exploitation \cite{agnihotri2020exploring}, and finally the acquisition function type (Upper Confidence Bounds ``UCB", Expected Improvement ``EI", Probability Of Improvement ``POI", etc). The sampler samples a new point ${\bm w}_1\sim\mathcal{N}(\mu,\Sigma)$ to form the auxiliary ellipse (lines 1-2). The known input point ${\bm w}_0$ is registered at $\theta=0,2\pi$ (lines 4-5). Now, $N_{\operatorname{BO}}$ samples are drawn using the acquisition function \cite{agnihotri2020exploring} which prioritizes samples from where the information gain is maximized (lines 9-10). We can stop after $N_{\operatorname{BO}}$ points (line 8) or once the uncertainty regarding the possible error in estimating the function $\varrho^\varphi\left(\theta\right)$ is under a threshold of our choice, and estimate the surrogate function using GP (line 13). We then find where the surrogate function intersects $\rho_{k-1}$ (line 15), and sample the new point using Eq.~\eqref{eqn:nonlin_ess_sample} (line 16). Fig.~\ref{fig:lin_nonlin_comp} depicts a snapshot of one ESS step for a linear system (Sec.~\ref{sec:demos_lin_ex_2dG}), for the purpose of comparisons. The surrogate function, black dash line, is estimated after sampling $N_{BO}$ red points. The confidence bounds are shown in turquoise and the segment that is in $\mathcal{L}_{{\bm w}}(\varphi, \rho_{k-1})$ is obtained in purple. The segment is correctly represented compared with the locations of the black triangles which are derived analytically using the closed-form solutions. Finally, the sample ${\bm w}$, in blue, is sampled uniformly from the purple active domain. 

\begin{algorithm}
\caption{Bayesian Optimization based sampler}\label{alg:bayes_sample}
\begin{algorithmic}[1]
\Require ${\bm w}_0$, $\rho_{k-1}, N_{\operatorname{BO}}, \kappa, \xi, \operatorname{type}=\{\operatorname{`UCB',`EI',`POI'}\}$
\State ${\bm w}_1 \sim \mathcal{P}({\bm w})$
\State $\varrho^\varphi(\theta) := \operatorname{Lambda}~\theta: \rho\left(r\left({\bm w}_0\cos(\theta)+{\bm w}_1\sin(\theta)\right)\right)$
\State $\operatorname{acq} \gets \operatorname{UtilityFunction}(\operatorname{type}, \kappa, \xi)$
\State $\operatorname{BO.Register}(0, \varrho^\varphi(0))$ \Comment{log the known input point}
\State $\operatorname{BO.Register}(2\pi, \varrho^\varphi(2\pi))$ 
\While{True}
\State $i \gets 0$
\While{$i \leq N_{\operatorname{BO}}$}
\State $\theta_i \gets \operatorname{BO.Suggest}(acq)$ 
    \State $\operatorname{BO.Register}(\theta_i, \varrho^\varphi(\theta_i))$ \Comment{log history}
\State $i \gets i+1$
\EndWhile
\State $\hat{\varrho}^\varphi,\hat{\varrho}^\varphi_\Sigma \gets \operatorname{BO.Predict}()$ \Comment{sample GP based on data}
\If{ $\operatorname{ValidPrediction}(\hat{\varrho}^\varphi,\hat{\varrho}^\varphi_\Sigma)$ }
    \State $\Theta \gets [\hat{\varrho}^\varphi \geq \rho_{k-1}]$ \Comment{the active set $\mathcal{L}$}
    \State $\theta \sim U[\Theta]$ \Comment{sample uniformly $\mathcal{L}$} 
    \State \Return ${\bm w}(\theta)$
\EndIf
\State $\operatorname{acq} \gets \operatorname{UtilityFunction}(\operatorname{type}, \kappa:=1, \xi:=0)$ ~~~~~~~~~~~~~~~~\Comment{gear towards exploitation mode}
\EndWhile
\end{algorithmic}
\end{algorithm}

At times, especially when reaching the smaller regions as $k$ approaches $K$, it might be hard to sample $\varrho^\varphi(\theta)\geq \rho_{k-1}$, and even though $\rho(r({\bm w}_0))=\varrho^\varphi(0)=\varrho^\varphi(2\pi)\geq \rho_{k-1}$ we might not be able to estimate where $\varrho^\varphi(\theta)\geq \rho_{k-1}$. If we exceed $N_{BO}$ samples, we propose to gear the BO towards exploitation (line 19)~\cite{agnihotri2020exploring} to add more samples and data next to $\theta=0,2\pi$, which are the current known maximum values.

\begin{figure}
    \centering
    \includegraphics[width=0.8\columnwidth]{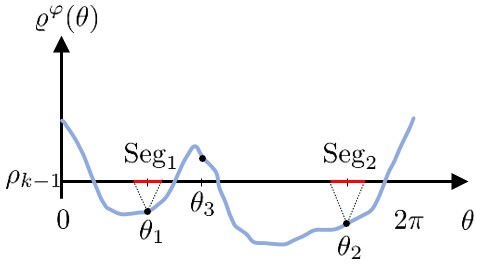}
    \caption{Lipschitz continuity-based sampler: $\theta_1$ is sampled from $U[0,2\pi]$, $\varrho^\varphi(\theta_1)<\rho_{k-1}$ so the red segment ($\text{Seg}_1$) is removed from $\Theta$. Same for $\theta_2$. The process ends here at $\theta_3$ where $\varrho^\varphi(\theta_3)>\rho_{k-1}$. The function represented with the blue line is not known.}
    \label{fig:lipschitz_sampler}
\end{figure}

(b) Lipschitz continuity-based elimination: the first technique presented is optimal, in the Bayesian sense, to estimate a surrogate function with minimal expensive simulation queries. However, it comes with a significant overhead to train and estimate the GP based on the history of samples, and to compute the acquisition function. Although it may be negligible compared to the time a single simulation takes, it accumulates and could still be considered expensive.

We propose Algorithm~\ref{alg:lipschitz_sample} that relies on the Lipschitz continuity theorem and based on Assumption \ref{assume_lipshitz}, that states that the function $\varrho^\varphi(\theta)$ cannot change more than the Lipschitz constant $M$, i.e. $|{d\varrho^\varphi(\theta)}/{d\theta}|\leq M$. The intuition is illustrated in Fig.~\ref{fig:lipschitz_sampler} and is based on the fact that sampling $\theta$ uniformly from $\Theta=[0,2\pi]$, which is then continuously being reduced, is equivalent to sampling uniformly from $\mathcal{L}(\varphi, \rho_{k-1})$. The process iterates until finding an ${\bm w}(\theta)\in\mathcal{L}_{\bm w}(\varphi, \rho_{k-1})$ (line 8). It starts by assigning the full domain $\Theta=[0,2\pi]$ (line 3) and samples a candidate $\theta_i\sim U[\Theta]$ (line 7). If ${\bm w}(\theta_i)\in\mathcal{L}_{\bm w}(\varphi, \rho_{k-1})$ we exit with ${\bm w}(\theta_i)$ (line 8). If it is less than the threshold, we discard the segment that is not plausible due to $M$, $\Theta = \Theta \setminus [\pm(\rho_{k-1}-\varrho^\varphi(\theta_i))/M+\theta_i]$ and continue on (line 10). In case we end up with $\Theta=\emptyset$, it means that our estimate for $M$ was too optimistic and we re-adjust $M=M\cdot(1+\epsilon)$, $\epsilon\in \mathbb{R}_+$ and re-evaluate the candidate domain $\Theta$ using the sampled history (lines 13-15). Doing this re-evaluation is computationally cheap because we do not use the simulator resources again. This mechanism allows us to provide a robust architecture while remaining parameter-free and requires no domain knowledge. 

\begin{algorithm}
\caption{Lipschitz continuity-based sampler}\label{alg:lipschitz_sample}
\begin{algorithmic}[1]
\Require ${\bm w}_0, M > 0$, $\rho_{k-1}$
\State ${\bm w}_1 \sim \mathcal{P}({\bm w})$
\State $\varrho^\varphi(\theta) := \operatorname{Lambda}~\theta: \rho\left(r\left({\bm w}_0\cos(\theta)+{\bm w}_1\sin(\theta)\right)\right)$
\State $\Theta \gets [0,2\pi]$
\State $\theta_\text{hist} \gets [~]$
\State $\rho_\text{hist} \gets [~]$
\While{True}
\State $\theta_i \sim U[\Theta]$ 
\If{$\varrho^\varphi(\theta_i) \geq \rho_{k-1}$}
    \Return ${\bm w}(\theta_i)$
\Else
    \State $\Theta \gets \Theta \setminus [\pm(\rho_{k-1}-\varrho^\varphi(\theta_i))/M+\theta_i]$
    \State $\theta_\text{hist} \gets \theta_\text{hist} \cup [\theta_i]$
    \State $\rho_\text{hist} \gets \rho_\text{hist} \cup [\varrho^\varphi(\theta_i)]$
    \If{$\Theta==\emptyset$}
        \State $M \gets M \cdot (1+\epsilon)$
        \State $\Theta \gets $ $\operatorname{ResetDomain}(\theta_\text{hist}, \rho_\text{hist}, M)$
    \EndIf
\EndIf
\EndWhile
\end{algorithmic}
\end{algorithm}

\subsubsection{Non-Gaussian uncertainties}
\label{sec:ver_bb_nf}
uncertainties originating from sensor noises, disturbances, and data-oriented algorithms (e.g. machine learning), cannot be usually modeled accurately with a unimodal Gaussian. For example, the Beam model \cite{thrun2002probabilistic} describes the error distribution of any range-measuring sensor. The model accounts for generic electronic white noise, multi-path errors, maximum range and translucent objects errors, or sensing other closer objects. These errors are multimodal and non-Gaussian. If we are to try to get a good estimate for $p^\varphi_f$, we must work with the model closest to the true distribution. Especially, when the true distribution is multimodal or heavy-tailed. Models can be obtained from data or derived mathematically, and are out of the scope of this work where we assume the distributions are known (Assumption \ref{assume_distribution}). 

Normalizing flows (NF) \cite{tabak2013family,kobyzev2020normalizing,papamakarios2021normalizing} is a general name for a collection of techniques used to create volume-preserving transformations between a usually simple and parametric base distribution to any other distribution, $\mathcal{P}$. In the recent past, it is commonly done using deep Neural Networks. We use NF to find a bijective function 
$b\left(\cdot\right)$ to transform a variable with a standard multivariate Normal distribution $X\sim\mathcal{N}\left(0, I\right)$, to produce a new transformed variable $W=b\left( X\right)$ where $W\sim\mathcal{P}$, and $b:\mathbb{R}^{l}\rightarrow\mathbb{R}^{l}$. So, to sample ${\bm w}\sim W$, we sample ${\bf x}\sim X$ and then compute ${\bm w}=b({\bf x})$. In our context, we use $b(\cdot)$ to transfer any sample ${\bf x}\sim\mathcal{N}\left(0, I\right)$ sampled in the STL-based ESS framework, to the corresponding uncertainty ${\bm w}\sim \mathcal{P}({\bm w})$ needed for the simulation. Another way to look at this is we are overloading another complexity upon the black-box simulator - the Gaussian uncertainty ${\bf x}$ is providing the necessary data for the simulator to build, with the use of $b(\cdot)$, the correct noise for its simulation. 

We must take special care with this approach because the probability to sample ${\bf x}$ and ${\bm w}$ might not be equivalent. This means that a point that is likely in $X$ might be ``over-sampled" in $W$ which may not accurately represent the probability density function (p.d.f) of $W$, and vice-versa. The bijective function introduces distribution \textit{warping}\cite{kobyzev2020normalizing}:
\begin{align}
\label{eqn:nf_density_corr}
    \log\left(\mathcal{P}_w\left({\bm w}\right)\right) = \log\left(\mathcal{P}_x\left(b^{-1}\left({\bm w}\right)\right)\right)-\log\left(\det \left|\frac{d{\bm w}}{d{\bf x}}\right|\right)
\end{align}
meaning, that the p.d.f at ${\bm w}$ is equal to the p.d.f of ${\bf x}$ plus a correction term that accounts for the transformation of $b(\cdot)$. We account for the warping by assigning weight $\alpha^k_r=\operatorname{det}|d{\bm w}/d{\bf x}|$ to every measurement $j=1,\ldots, N_{\operatorname{ESS}}$ taken with STL-based ESS when computing the conditional probability of nesting $k$. 
\begin{align}
    p(\mathcal{L}_{k}|\mathcal{L}_{k-1}) = \frac{ \sum_{j=1}^{N_{\operatorname{ESS}}} \alpha^{k-1}_j {\bf 1}\left(x_j\in \mathcal{L}_k\right)}{\sum_{j=1}^{N_{\operatorname{ESS}}} \alpha^{k-1}_j {\bf 1}\left(x_j\in \mathcal{L}_{k-1}\right)}
\end{align}
With performance borne in mind, the process of learning the function $b(\cdot)$ is done only once and is used throughout the STL-based ESS process. The function can be concatenated in a chain with other functions to estimate complicated distributions. In our case studies we use \texttt{Pyro}~\cite{bingham2018pyro} to learn the transformation.

\subsection{Error Analysis}
\label{sec:verification_error_analysis}

\noindent Estimating the probability of failure $p_f^\varphi$ using an MCMC sampler is by definition stochastic. Meaning, the estimate comes with an inherent error. In this section, we derive a quantitative measure of confidence on the probability. We differentiate between the error sources in the linear and non-linear formulations.

\subsubsection{Linear STL-based ESS}
\label{sec:linear_stl_ess_error_analysis}
the probability is computed as the product of conditional probabilities of each of the nestings, Eq.~\eqref{eqn:p_prod_cond}. The ESS process in every nesting is MCMC, namely, a new sample depends on previous samples (the Markov chain). However, we break this dependency with the ``burn-in" phase, by discarding $N_{\text{skip}}$ samples between two accepted samples. The burn-in sampling and the uniformly sampled $\theta$ from the active domain, all weaken the dependencies to make it (nearly) iid. With this assumption, we can apply the Central Limit Theorem (CLT), given $N_{\text{ESS}}\cdot p(\mathcal{L}_k|\mathcal{L}_{k-1})\gg 1$. We will denote the conditional probability of nesting $k$ given $k-1$ as $p_k$ for brevity. Since we design the process such that each $p_k\approx 0.5$, and the variance of the probability given $N_{\text{ESS}}$ samples, is $\operatorname{Var}_k \approx p_k(1-p_k)/N_{\text{ESS}}$, the distribution of the conditional probability is $p_k  \sim \mathcal{N}(\tfrac{1}{2}, \tfrac{1}{4N_{\text{ESS}}})$ and each conditional probability $p_k$ is iid. As a result, we can compute the variance of the product of the conditionals \cite{scher2022elliptical}:

\begin{flalign}
\label{eqn:var_of_nestings}
\operatorname{Var}[p_1\cdots p_K] 
&= \EX[(p_1\cdots p_K)^2]-\left(\EX[p_1\cdots p_K]\right)^2\\\nonumber
&= \EX[p_1^2\cdots p_K^2]-\left(\EX[p_1]\cdots \EX[p_K]\right)^2\\\nonumber
&= \EX[p_1^2]\cdots \EX[p_K^2] - (\EX[p_1])^2\cdots (\EX[p_K])^2\\\nonumber
&= \prod_{k=1}^K \left(\operatorname{Var}[p_k]+(\EX[p_k])^2\right)
- \prod_{k=1}^K \left(\EX[p_k]\right)^2
\end{flalign}
We get an approximation by substituting for our nominal parameters: 
\begin{flalign}
\label{eqn:hdr_approx_var}
\operatorname{Var}[p_1\cdots p_K] = \prod_{k=1}^K \left(\frac{1}{4N_{\text{ESS}}}+\frac{1}{4}\right)
- \prod_{k=1}^K \left(\frac{1}{4}\right)
\end{flalign}
Using this approximation, one may also derive the number of samples per nesting needed to obtain an adequate level of confidence. See sections 4.4.2-3 in \cite{scher2022elliptical} for more details and a plot of the confidence level versus $N_{\text{ESS}}$ and $K$.

\subsubsection{Precise sampling - Guaranteeing performance}
while the results in \eqref{eqn:var_of_nestings} are accurate, we have no way to guarantee a-priori what will be the maximum error we can provide. The reason is that we do not know what the variance of $p_k^2$ could be, $\operatorname{Var}[p_k^2]=\operatorname{Var}[p_k]+\mathbb{E}[p_k]^2$. In this section, we provide maximum error guarantees with a confidence measure. This could be used by the designers and regulator as a safety margin on the provided failure probabilities.

\begin{figure}
     \centering
     \includegraphics[width=\columnwidth,height=4cm,keepaspectratio]{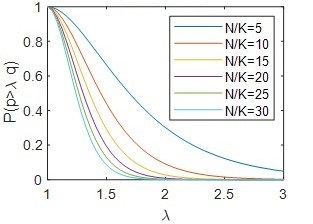}
     \caption{The probability of estimating $p_f^\varphi$ that is greater than $\lambda$ times the true probability $q$, as a function of $N_{\operatorname{ESS}}/K$.}
     \label{fig:prob_err_bounds}
\end{figure}

\textit{Error upper bound}: Let $q_1,\ldots,q_K$ be the true and latent probabilities of nestings $k=1,\ldots,K$ and also assume $q_k\geq \delta> 0$, i.e. $q_k \in [\delta,1]$. Let $p_1,\ldots,p_K$ be the random variables that estimate the true nesting probabilities respectively.  $p_k$ is an unbiased estimator for $q_k$, i.e. $\mathbb{E}[p_k]=q_k$, where
\begin{flalign}
p_k = \frac{1}{N_{\operatorname{ESS}}} \sum_{j=1}^{N_{\operatorname{ESS}}} {\bf 1}\left(x_{k,j}\sim\mathcal{L}_{k-1} \in \mathcal{L}_{k}\right),
\end{flalign}
where $\mathbb{P}(x_{k,j}\sim\mathcal{L}_{k-1} \in \mathcal{L}_{k} )=1-\mathbb{P}(x_{k,j}\sim\mathcal{L}_{k-1} \notin \mathcal{L}_{k} )=q_k$ and we assume that all $x_{k,j}$'s are independent (according to Sec.~\ref{sec:linear_stl_ess_error_analysis}) and $p_k$ follows the Binomial distribution (B3). We create an auxiliary unbiased random variable, $Y_{k,j}=({\bf 1}\left(x_{k,j}\sim\mathcal{L}_{k-1}\in \mathcal{L}_{k}\right)-q_k)/q_k$ thus $Y_{k,j}\in [-1, (1-q_k)/q_k]$ and $\mathbb{E}[Y_{k,j}]=0$.  
\begin{flalign}
    \frac{p_k-q_k}{q_k}=&\frac{1}{Nq_k}\sum_{j=1}^{N_{\operatorname{ESS}}}\left( {\bf 1}\left(x_{k,j}\sim\mathcal{L}_{k-1}\in \mathcal{L}_{k}\right) - q_k\right) \\\nonumber =& \frac{1}{N_{\operatorname{ESS}}}\sum_{j=1}^{N_{\operatorname{ESS}}} Y_{k,j}
\end{flalign}
and the total probability $p$ is the product of all the conditional probabilities from each nesting in the HDR process. We search for a bound on the multiplicative error from the true probability, $q$: 
\begin{flalign}
\label{eqn:p_over_q_bound}
    \frac{p}{q} = &\prod_{k=1}^K \frac{p_k}{q_k}=\prod_{k=1}^K\left( 1+\frac{p_k-q_k}{q_k}\right) \leq \\\nonumber &\exp\left( \frac{1}{N_{\operatorname{ESS}}}\sum_{k=1}^K\sum_{j=1}^{N_{\operatorname{ESS}}} Y_{k,j}\right)
\end{flalign}
using that for all real $z$, $1+z\leq e^z$ and that $\prod_i (1+z_i)\leq \prod e^{z_{i}} = \exp(\sum_i z_i)$.
For any $\epsilon>0$ we have by Hoeffding's inequality (B2) that $\mathbb{E}[e^{\epsilon Y_j}]\leq e^{\epsilon^2(b-a)^2/8}$. Using $Y_{k,j}$ bounds,
\begin{flalign}
    \mathbb{E}\left[ e^{\epsilon Y_{k,j}}\right] \leq e^{\sfrac{\epsilon^2}{8q_j^2} } \leq e^{\sfrac{\epsilon^2}{8\delta^2} }
\end{flalign}
Thus, for any scalar $M>0$,
\begin{flalign}
    &\mathbb{P}\left[ \frac{p}{q} \geq e^M \right] \leq \mathbb{P}\left[\sum_{k=1}^K\sum_{j=1}^{N_{\operatorname{ESS}}} Y_{k,j} \geq N_{\operatorname{ESS}}M\right] \\\nonumber
    &\leq \prod_{k=1}^K\prod_{j=1}^{N_{\operatorname{ESS}}} \mathbb{E}\left[ e^{\epsilon Y_{k,j}}\right] e^{-\epsilon N_{\operatorname{ESS}}M} \leq e^{\frac{KN_{\operatorname{ESS}}}{8\delta^2}\epsilon^2 - N_{\operatorname{ESS}}M\epsilon}
\end{flalign}
where the first inequality comes from the bound in \eqref{eqn:p_over_q_bound}, the second inequality is the consequence of the Markov inequality (B1). Minimizing over $\epsilon>0$, we choose $\epsilon=4\delta^2M/K$ to get:
\begin{flalign}
    \label{eqn:mult_bound}
    \mathbb{P}(p \geq \lambda q) \leq e^{-2\delta^2(\log(\lambda))^2N_{\operatorname{ESS}}/K }
\end{flalign}
For example, assuming that $\delta=0.5$ and $K=10, N_{\operatorname{ESS}}=250$ yields that the probability that our estimate is greater than twice the true probability, i.e. $\lambda=2$, is bounded by $\mathbb{P}\leq 0.0025$. Selecting $\lambda=2.5$ with the same parameters, bounds the probability at $\mathbb{P}\leq 0.000028$. Of course, \eqref{eqn:mult_bound} can also be used to make an informed decision about the number of samples to take per nesting, as shown with Fig.~\ref{fig:prob_err_bounds}. The scalar $\delta$ is not known, but can be assumed or approximated using the subset part of the algorithm \cite{gessner2020integrals}. It is crucial that $\delta=\min q_k$ is not too low.

\section{Synthesis}
\label{sec:synthesis_intro}

\noindent The method introduced in Sec.~\ref{sec:verification_intro} enables us to sample from specification-violating trajectories $\mathcal{L}(\varphi,0)$ efficiently once we have completed evaluating $p_f^\varphi$ and have one or more violating examples. As we will see in this section, we can utilize this sampling ability to efficiently obtain the gradients of the probability w.r.t any parameter of the system and use gradient descent to find a new parameter that drives the probability to fail down. We will be able to synthesize new controls or test out parameters of the system, objective 2, which mean a safer operation of the robot, at least locally (i.e. not a global optimum). Unlike the other techniques presented in Sec.~\ref{sec:relwork}, which guarantee the best performance or best robustness only for the nominal execution, the advantage of this method is the guarantee of the best safety for all possible outcomes (minimize $p_f^\varphi$). 

The design pipeline is as follows. Given a system, first, we must obtain the controls and commands that satisfy, or occasionally satisfy, the specification. This can be achieved in many different ways, for example, way-points navigation, RRT~\cite{lavalle1998rapidly}, non-linear trajectory optimization, control barrier functions (CBF) or Mixed Integer Programming (MIP), as discussed in Sec.~\ref{sec:relwork}. In step two, we derive a trajectory-stable controller using, for example, Linear Quadratic Regulator (LQR)~\cite{kwakernaak1972linear}. We then loop until $p_f^\varphi$ reaches a minimum or the desired performance level. At each iteration $i$, we compute $p_f^\varphi$ with the current parameter $\gamma_i$, find the derivatives w.r.t the parameter of choice $\gamma_i$, and re-evaluate $\gamma_{i+1}$ using gradient descent, Eq.~\eqref{eqn:synth_gradient_descent}.  

\subsection{Linear Systems with Gaussian Uncertainties}
\label{sec:synthesis_lingauss}

\noindent As portrayed in \cite{gessner2020integrals}, Eq.~\eqref{eqn:eq_stl_integral} can be written more explicitly:
\begin{align}
    \label{eqn:synth_integral}
    p_f^\varphi=&p({\bf x}_{traj} \in \mathcal{L}(\varphi,0))=\mathbb{E}\left[X\in\mathcal{L}(\varphi,0)\right]= \\\nonumber &\underset{\mathbb{R}^{n\cdot N}}{\int} \prod_{j\in [1...N_\mu]} {\bm 1}(H'_j{\bf x}+h_j) ~\mathcal{N}({\bf x};\mu,\Sigma) d{\bf x},
\end{align}
where $\mathcal{N}({\bf x};\mu,\Sigma)$ means that the latent variable ${\bf x}\sim\mathcal{N}(\mu, \Sigma)$ for $N_\mu$ trajectory-level predicates.
Let $\lambda = \{\mu,\Sigma\}$, we can derive $p_f^\varphi$:
\begin{align}
    \label{eqn:synth_derivation}
    \frac{dp_f^\varphi}{d\lambda}= \underset{\mathbb{R}^{n\cdot N}}{\int} d{\bf x}\prod_{j\in [1...N_\mu]} {\bm 1}(H'_j{\bf x}+h_j) \frac{d\mathcal{N}({\bf x};\mu,\Sigma)}{d\lambda},
\end{align}
The multivariate Gaussian p.d.f is defined as:
\begin{align}
    \label{eqn:gaussian_pdf}
\mathcal{N}({\bf x};\mu,\Sigma) = &(2\pi)^{-nH^\varphi/2}|\Sigma|^{-1/2}\cdot \\\nonumber & \exp\left(-0.5({\bf x}-\mu)'\Sigma^{-1}({\bf x}-\mu)\right)
\end{align}
for ${\bf x}\in\mathbb{R}^{n\cdot N}$, so the following expression is also true:
\begin{align}
    \label{eqn:synth_helpereqn}
    \frac{d\mathcal{N}({\bf x};\mu,\Sigma)}{d\lambda}=\mathcal{N}({\bf x};\mu,\Sigma)\frac{d\log\mathcal{N}({\bf x};\mu,\Sigma)}{d\lambda} ,
\end{align}
Then, plugging Eq.~\eqref{eqn:synth_helpereqn} in Eq.~\eqref{eqn:synth_derivation} reveals that:
\begin{align}
    \label{eqn:synth_lambdagradient}
    \frac{dp_f^\varphi}{d\lambda}= \mathbb{E}\left[\frac{d\log\mathcal{N}({\bf x};\mu,\Sigma)}{d\lambda}\right],
\end{align}
Eq.~\eqref{eqn:synth_lambdagradient} implies that after evaluating $p_f^\varphi$, we can obtain $N_{\operatorname{samples}}$ \emph{new} samples from $\mathcal{L}(\varphi,0)-$ and compute the gradient with the expected value of Eq.~\eqref{eqn:synth_lambdagradient}. As shown, ESS allows us to sample rejection-free thus it is efficient. The number of samples $N_{\operatorname{samples}}$ can be significantly less than the number of samples needed with finite differences, $2\times n\times N$ and computing $p^\varphi_f$ for each one of them. The expression in Eq.~\eqref{eqn:synth_lambdagradient} can be further simplified by evaluating the derivatives:
\begin{align}
    \label{eq:synth_explicit_deriv}
    & \frac{d\log\mathcal{N}({\bf x};\mu,\Sigma)}{d\mu} = \Sigma^{-1}({\bf x}-\mu) \\ \nonumber 
    & \frac{d\log\mathcal{N}({\bf x};\mu,\Sigma)}{d\Sigma} = 0.5\left(\Sigma^{-1}({\bf x}-\mu)({\bf x}-\mu)'\Sigma^{-1}-\Sigma^{-1}\right)
\end{align}

The chain rule is used when we want to find the gradient w.r.t another parameter in the system, for example, the controller $K$. Bear in mind that a parameter can influence the trajectory both in $\mu, \Sigma$ or separately, recall Eq.~\eqref{eqn:trajectory_pdf}. 

\begin{example}
We present the gradients for the reference trajectory ${\bm r}_{ref}$ and controller $K$, Eq.~\eqref{eqn:synth_grad_by_uref}-\eqref{eqn:synth_grad_by_k} respectively:
\begin{align}
    \label{eqn:synth_grad_by_uref}
    \frac{dp_f^\varphi}{dr_{ref}} = \frac{dp_f^\varphi}{d\mu}\frac{d\mu}{dr_{ref}} = 
    \Phi_r^T\mathbb{E}\left[ \Sigma^{-1}({\bf x}-\mu)\right]   
\end{align}
\begin{align}
    \label{eqn:synth_grad_by_k}
    &\frac{dp_f^\varphi}{dK}_{ij} = \frac{dp_f^\varphi}{d\mu}\frac{d\mu}{dK} + \frac{dp_f^\varphi}{d\Sigma}\frac{d\Sigma}{dK} = \sum_{k=1}^{T} \frac{d\mu_k}{dK_{ij}}\mathbb{E}\left[ \Sigma^{-1}({\bf x}-\mu)\right]_k \\ \nonumber 
    &+\sum_{k=1}^{T}\sum_{l=1}^{T} \frac{d\Sigma_{kl}}{dK_{ij}}\mathbb{E}\Big[ 0.5\big(\Sigma^{-1}({\bf x}-\mu)({\bf x}-\mu)'\Sigma^{-1} -\Sigma^{-1}\big)\Big]_{kl} 
\end{align}
\end{example}
To perform the gradient descent step, we need a learning rate, $\alpha$. However, when optimizing for cases where $p_f^\varphi\ll 1$, we want smaller steps so we do not overshoot. We introduce $p_f^\varphi$ in the gradient descent update as a normalizing factor so that the learning rate is easier to obtain and can remain relatively constant. We also introduce a sign variable where $v_{dir}=1$ if we search for $p_s^\varphi$ and $v_{dir}=-1$, if we search for $p_f^\varphi$. The gradient descent process for some parameter $\gamma$ of the system, is then:
\begin{align}
\label{eqn:synth_gradient_descent}
    \gamma_{i+1} = \gamma_i + v_{dir}\alpha \frac{dp_f^\varphi}{d\gamma_i} p_f^\varphi
\end{align}
 
\noindent Some notes regarding the synthesis: 

\noindent (a) It is possible to derive the Hessian as well (see~\cite{gessner2020integrals}). A possible improvement to the gradient descent is by incorporating the second-degree derivative for more accuracy.

\noindent (b) The lower the probability of failure is, the longer the computation (more nestings). Keep in mind that $p_f^\varphi=0$ is not possible due to the assumption that the uncertainty is an unbounded Gaussian. Therefore, it is up to the designer to specify the stopping criteria at the requisite level of performance.

\noindent (c) STL-based ESS can also attempt to find the initial trajectory. We can exploit the control signal $u_t\sim\mathcal{N}(0,\Sigma^u_{t})$ as the ``uncertainty''. Then, STL-based ESS can be performed until a control sequence $u_0,\ldots,u_{N-1}$ that satisfies the specification is found, even if it is a very low probability. The variance in $\Sigma^u_{t}$ is used to attenuate the ``bandwidth" of the control signal. Note that other works, proposed in the preface of this section, are usually faster and better suited to find nominal starting trajectories and controls. 

\subsection{Black-box Systems with non-Gaussian Uncertainties}
\label{sec:synthesis_blackbox}

\noindent Estimating $p_f^\varphi$ in the non-linear verification problem is equivalent to solving the integral  (Sec.~\ref{sec:verification_blackbox}):
\begin{align}
p_f^\varphi &= \int_{{\bm w}\in \mathbb{R}^l} \prod_{d=[1..N_\mu]} \Theta\left(\rho\left(r\left({\bm w,\gamma}\right)\right)\right)\mathcal{N}\left({\bm w};\mu,\Sigma\right)d{\bm w} \\ \nonumber
&= \int_{{\bm w}\in \mathbb{R}^l} \Psi\left({\bm w,\gamma}\right)\mathcal{N}\left({\bm w};\mu,\Sigma\right)d{\bm w}
\end{align}
where the function $\Psi$ added for clarity, represents $1$ where ${\bm w}\in\varphi$ and $0$ otherwise. Note that this is  a substantially different problem than the linear systems problem, in  Eq.~\eqref{eqn:synth_integral}. In Sec.~\ref{sec:synthesis_lingauss}, $\mu$ and $\Sigma$ represent the trajectory (state over time) distribution, and are affected by the change in the parameter $\gamma$. The distribution changes, but the predicates from the specification remain the same. In the non-linear and black-box systems, ${\bm w}\sim\mathcal{N}\left(\mu,\Sigma\right)$ is the input, it is given and thus $\mu$ and $\Sigma$ do not depend on $\gamma$. The distribution is unchanging. On the other hand, the domains where the specification holds, change with changes in $\gamma$. Using the Leibniz integral rule:
\begin{align}
\frac{dp}{d\gamma} &= \int_{{\bm w}\in \mathbb{R}^l} \frac{d}{d\gamma}\Psi\left({\bm w,\gamma}\right) \mathcal{N}\left({\bm w};\mu,\Sigma\right) d{\bm w}
\end{align}
A small change in $\gamma$ may shift the domains completely and it is not apparent how to find that shift efficiently where there is no information (black box) or reasonable assumptions on these changes. We conclude that it is not trivial to obtain the gradient of $p_f^\varphi$ in an efficient and accurate manner with black-box systems in this method and will be further investigated in future work.
\section{Case Studies}
\label{sec:demos}

\noindent With the exception of example A, we present test-cases with probabilities that are not low enough to be considered rare, so we can compare with MC.

\subsection{The probability of a constrained 2D Gaussian}
\label{sec:demos_lin_ex_2dG}
\noindent We start with a synthetic example from  \cite{sinha2020neural} where the uncertainty ${\bm w}\sim\mathcal{N}(0,\text{I})\in\mathbb{R}^2$ and the goal is to find the probability to sample where $\min(|w_1|,w_2)\geq -\beta$. To translate it to our formalism, the system and specifications are,
\begin{align}
\label{eqn:prbo_2d_sys_eq}
    {\bf x} &= {\bm w} \\ \nonumber
    \varphi_\beta &:= (x_1\geq -\beta \wedge x_2\geq -\beta) \vee (-x_1\geq -\beta \wedge x_2\geq -\beta)
\end{align}
This problem is chosen both because: (a) it is relatively challenging where the probability is low, yet there are two distinct regions equally distant from the origin and we must capture both; (b) because it has an analytical solution, $p=2(1-0.5(1 + \operatorname{erf}(-\beta/\sqrt{2})))^2$ where $\operatorname{erf}$ is the error function\footnote{\url{https://en.wikipedia.org/wiki/Error_function}}, so we can evaluate the performance w.r.t the ground truth.

\subsubsection{Verification} 
\begin{figure}
     \centering
     \includegraphics[width=\columnwidth,height=4cm,keepaspectratio]{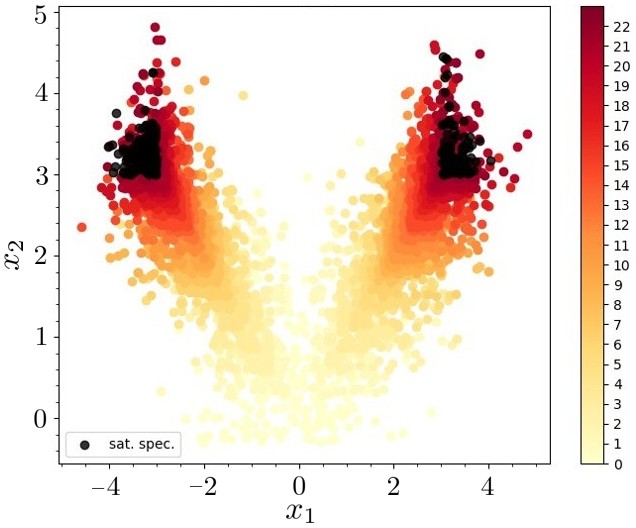}
     \caption{The sampling process of STL-based ESS for the 2D Gaussian problem for $\beta=-3$. Black samples are sampled from $\mathcal{L}(\varphi_\beta,0)$ and the lighter the color, it was sampled in an earlier nesting.}
     \label{fig:twod_ex}
\end{figure}

Fig.~\ref{fig:twod_ex} shows the process and advancement of sampling in the different nestings for $\beta=-3$, where $p=3.6\cdot10^{-6}$. The black samples represent samples from the final domain $\mathcal{L}(\varphi,0)$. We can see how the ESS and HDR propagate the samples in the direction of the satisfying regions rather than sample uniformly and inefficiently in ${\bm w}$ as MC does.

\begin{figure}
     \centering
     \includegraphics[width=\columnwidth,height=4cm,keepaspectratio]{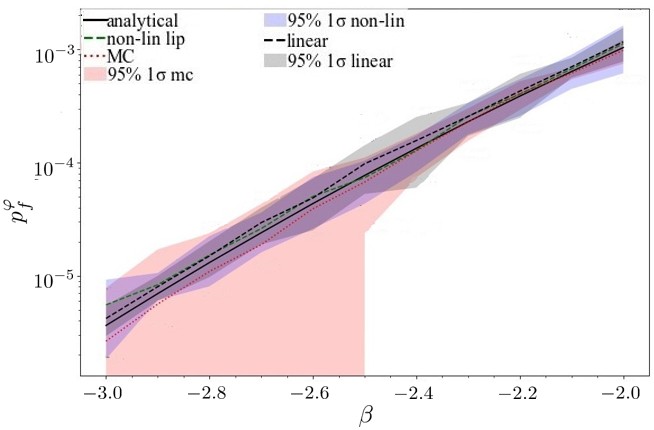}
     \caption{Computing $p_s^\varphi$ as a function of $\beta$ and comparison of STL-based ESS between linear, non-linear on MC. At lower probabilities, the number of samples does not allow MC to be confident and some experiments end with $0$ satisfying examples. }
     \label{fig:twod_results}
\end{figure}

We now evaluate the performance for different values of $\beta\in[-3,\ldots,-2]$. We run each $\beta$-experiment $10$ times to obtain an estimate of the probability mean and variance. We compare the results of the analytic solution to running STL-based ESS for the linear formulation, the non-linear formulation, and MC in Fig.~\ref{fig:twod_results}. The number of simulations in MC is chosen as the same number needed for the non-linear STL-based ESS. We can see how the variance of MC is very high when $\beta<-2.5$ (1-standard deviation indicated with the shaded area) and the probabilities are low. The performance, in terms of the number of simulations versus accuracy, is comparable to \cite{sinha2020neural}. Better performance or faster-obtained results may be achieved by increasing $N_{\operatorname{ESS}}$ and $N_{\operatorname{skip}}$ or decreasing, respectively.

\subsubsection{Synthesis} 
\begin{figure}
     \centering
     \begin{subfigure}[b]{\columnwidth}
         \centering
         \includegraphics[width=6cm,keepaspectratio]{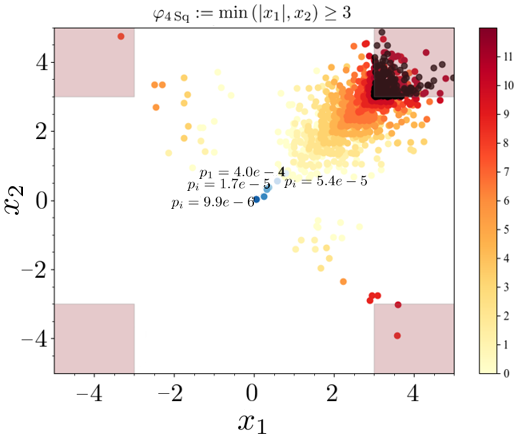}
         \caption{}
         \label{fig:2d4d_initial}
     \end{subfigure}
     \vfill
     \begin{subfigure}[b]{\columnwidth}
         \centering
         \includegraphics[width=6cm,keepaspectratio]{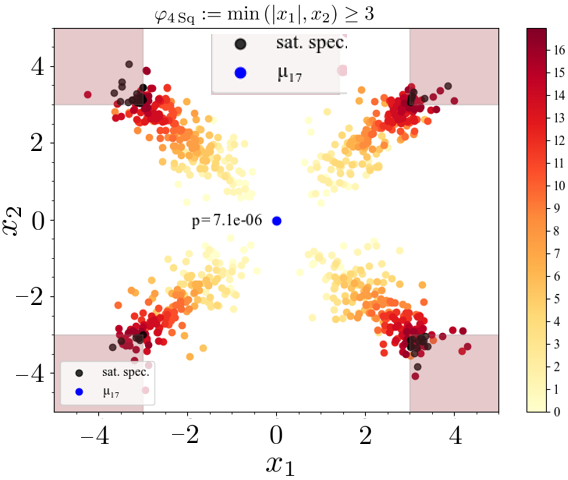}
         \caption{}
         \label{fig:2d4d_final}
     \end{subfigure}
        \caption{ Synthesis process for the 2D Gaussian example with 4 domains, $\varphi_{4\text{sq}}$. (a) We start with $\mu_0=[1,1]'$. Note that STL-based ESS is sampling mostly from the top-right domain since it is closest. The blue dots represent the progress of $\mu_i, i=[0...17]$. (b) The STL-based ESS process for the last iteration, sampling from all $4$ domains where $p=7.2e-6$ is the ground truth. }
        \label{fig:2d_4d}
\end{figure}

we demonstrate synthesis in this example to show how to find the parameters that minimize $p$. Suppose the new specification requires four domains of interest:
\begin{align}
\label{eqn:prbo_2d_sys_eq_4squares}
    \varphi_{4\text{sq}} := &(x_1\geq 3 \wedge x_2\geq 3) \vee (-x_1\geq 3 \wedge x_2\geq 3) \vee \\\nonumber
    & (x_1\geq 3 \wedge -x_2\geq 3) \vee (-x_1\geq 3 \wedge -x_2\geq 3)
\end{align}
It is easy to verify that given a fixed $\Sigma$, a globally optimal value for $\mu$ is $[0,0]'$ (see Fig.~\ref{fig:2d4d_final}). We begin with $\mu_0=[1,1]'$ w.l.o.g. and $\Sigma=\text{I}_2$. We implement the gradient descent from Sec.~\ref{sec:synthesis_lingauss} with a step size of $\alpha=0.1$ to find $\mu$ that minimizes the probability to sample from any of the four regions. After $18$ iterations illustrated in Fig.~\ref{fig:2d4d_initial}, $\mu_{18}=[0.04,0.04]'$. 

\subsection{Autonomous Vehicle with perception noise}
\label{sec:demos_autonomouscar}

\begin{figure}
     \centering
     \begin{subfigure}[b]{0.49\columnwidth}
         \centering
         \includegraphics[width=\columnwidth]{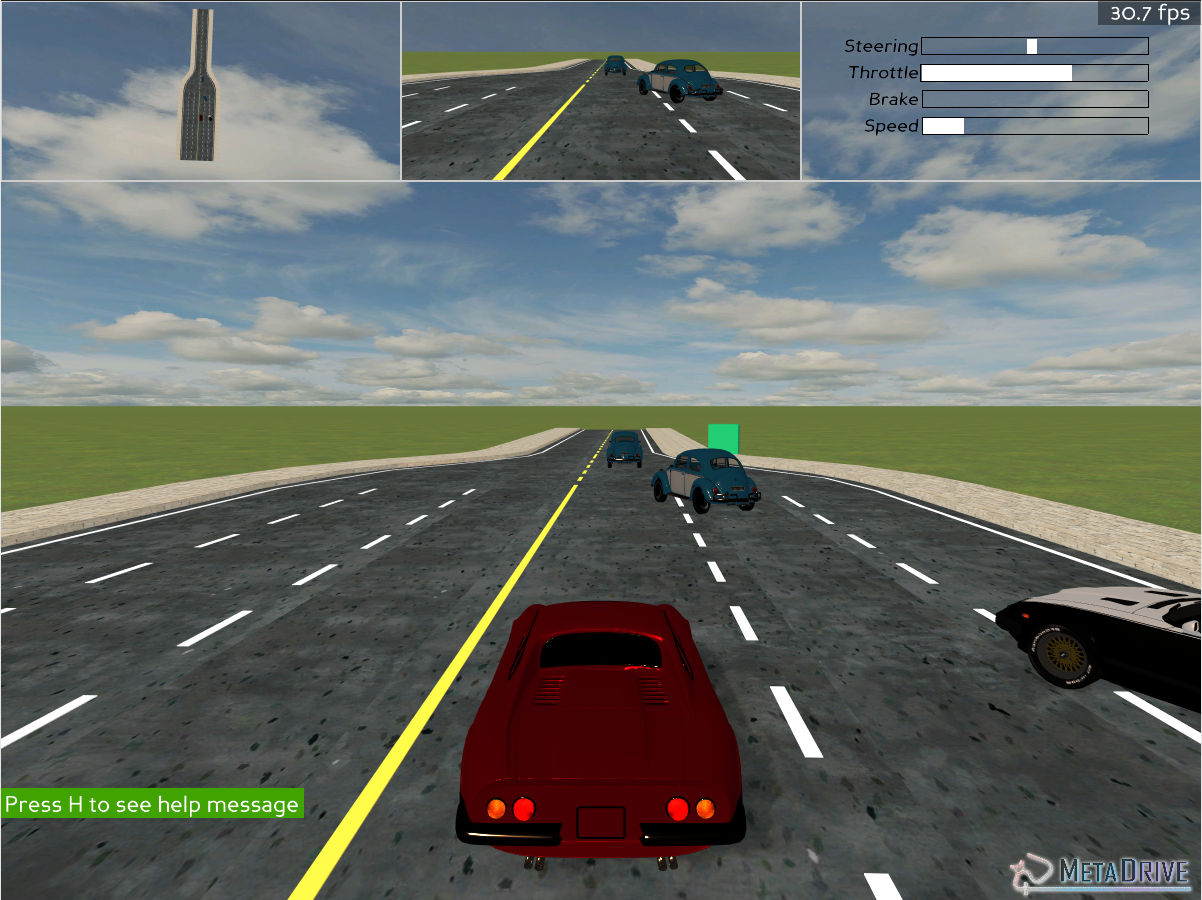}
         \caption{}
         \label{fig:av_pv}
     \end{subfigure}
     \hfill
     \begin{subfigure}[b]{0.38\columnwidth}
         \centering
         \includegraphics[width=\columnwidth, height=3.3cm  ]{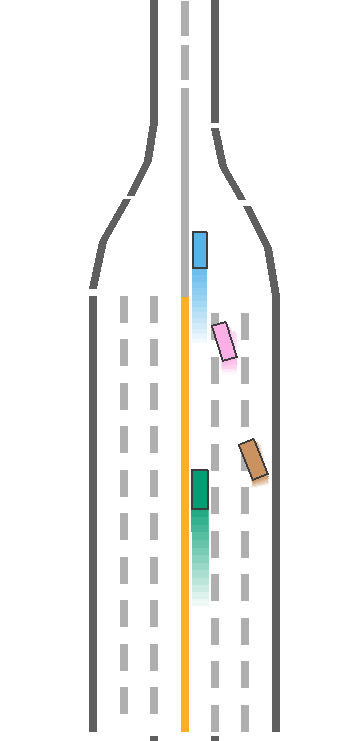}
         \caption{}
         \label{fig:av_ev}
     \end{subfigure}
        \caption{\texttt{Metadrive} simulator \cite{li2021metadrive}. (a) The view behind the (red) ego car and the view from the dashboard camera, on the top. (b) A top-view of the road and traffic, ego car in green.  }
        \label{fig:av_simulation}
\end{figure}

\noindent In this example, an autonomous car is trained to drive and avoid traffic using a 240-ray Lidar sensor plus several more features like the lateral location in the lane, speed, etc, using deep reinforcement learning, see the environment in Fig.~\ref{fig:av_simulation}. The simulation environment \texttt{Metadrive}~\cite{li2021metadrive}, is a realistic physics-based driving simulator. In operational time, the measurements are fed to a neural network that produces deterministic steering and throttle commands. The goal is to reach a certain reward and to avoid slowing down under $V_m=18\text{m/s}$ for more than $5$ consecutive steps in a road scenario of merging traffic lanes:
\begin{flalign}
\label{eqn:spec_autonomous_car}
    \varphi_{AV} :=& \Box_{[30,200]} \left(\text{Speed} < V_m \implies  \Diamond_{[0,4]} \left(\text{Speed} \geq V_m\right)\right)   \\ \nonumber
    \wedge & \Diamond_{[0,200]} \left(\text{Reward}\geq 207 \right)
\end{flalign}
Both measurements may be indicators of a crash that occurred, but they may also happen regardless, due to congestion.
The noise dimensionality is thus $l=200\cdot240=48\text{k}$, with an added truncated Normal distribution $w_{i,t}
\sim\mathcal{N}\left(0,0.004\right), \forall i\in[1,240], t\in[0,200]$ at the lidar's maximum (and minimum) range. The noise on the readings may cause phantom objects in the perception, which may cause the car to slow down or hit another car and violate $\varphi_{AV}$. The output ${\bf y}$ has the vehicle's speed, steering, yaw rate, lateral left and right position in the lane, number of crashes, number of times it went out of the road, and cumulative reward and the instantaneous reward at every time step.

\subsubsection{Results} 
\begin{figure}
     \centering
     \begin{subfigure}[b]{0.49\columnwidth}
         \centering
         \includegraphics[width=\columnwidth]{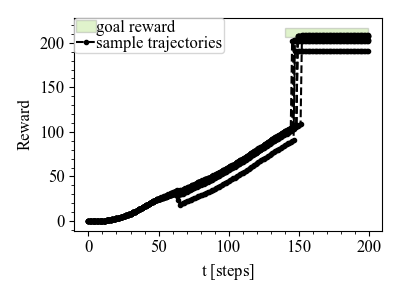}
         \caption{Cumulative reward}
         \label{fig:av_res_reward}
     \end{subfigure}
     \hfill
     \begin{subfigure}[b]{0.49\columnwidth}
         \centering
         \includegraphics[width=\columnwidth]{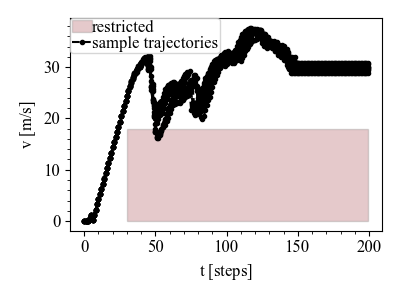}
         \caption{Vehicle speed}
         \label{fig:av_res_speed}
     \end{subfigure}
     \vfill
     \centering
     \begin{subfigure}[b]{0.49\columnwidth}
         \centering
         \includegraphics[width=\columnwidth]{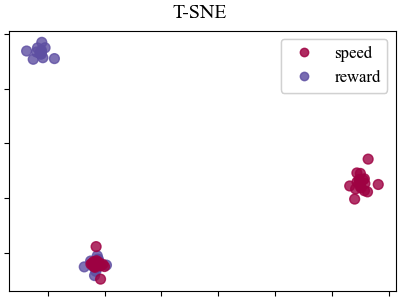}
         \caption{High-dimensional noise embedding with T-SNE}
         \label{fig:av_tsne}
     \end{subfigure}
        \caption{Failing perching trajectories}
        \label{fig:av_results}
\end{figure}

sampled failing trajectories are shown in Fig.~\ref{fig:av_results}. The probability to fail according to our method $p_f^\varphi=0.5\%$ using $13449$ simulations ($p_f^\varphi=0.4\pm0.05\%$ with MC and the same number of simulations). Some trajectories fail for not reaching the reward goal, mainly due to a collision that occurred during the run, Fig.~\ref{fig:av_res_reward}. Some trajectories reached the reward goal however they slowed down too much instead of taking over another car, Fig.~\ref{fig:av_res_speed}. Further analysis with T-SNE \cite{van2008visualizing} of the noise causing the failing trajectories reveals that the noises are clustered in three distinct groups. One group causes the reward failure (collision), one group causes the speed to drop, and one fails both, in Fig.~\ref{fig:av_tsne}.

\subsection{A perching fixed-wing plane}
\label{sec:demos_perching}

\begin{figure}
     \centering
     \includegraphics[width=\columnwidth]{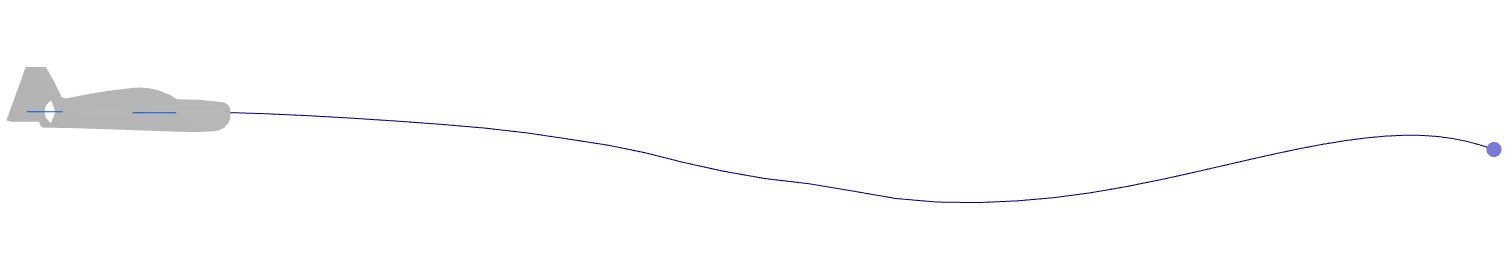}
     \caption{Fixed-wing perching and the trajectory planning.}
     \label{fig:perching_view}
\end{figure}

\noindent The authors in \cite{moore2012control, moore2014robust} developed a time-varying LQR controller and an optimal trajectory to land a fixed-wing plane on a wire (Fig.~\ref{fig:perching_view}) in highly non-linear dynamics, similar to how birds approach a branch for perching. The development generally assumed perfect knowledge of the states and here we evaluate the robustness to noise. We have several options for the introduction of noises and we focus on two: actuator noise at each time-step; and the initial location of the plane. 

\subsubsection{Initial location} 
\begin{figure}
     \centering
     \begin{subfigure}[b]{0.49\columnwidth}
         \centering
         \includegraphics[width=\columnwidth]{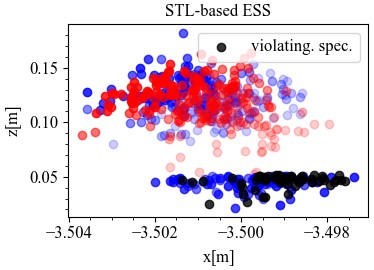}
         \caption{STL-based ESS}
         \label{fig:perching_ver_ic_ess}
     \end{subfigure}
     \hfill
     \begin{subfigure}[b]{0.49\columnwidth}
         \centering
         \includegraphics[width=\columnwidth]{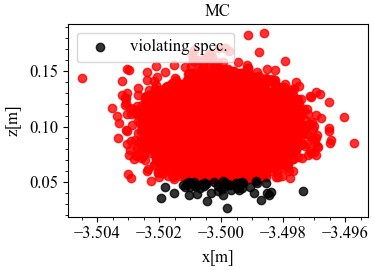}
         \caption{MC}
         \label{fig:perching_ver_ic_mc}
     \end{subfigure}
        \caption{Perching verification of noisy initial conditions with $8,880$ simulations: (a) $p_f^{\varphi_p}=0.85\%$. (b)  $p_f^{\varphi_p}=0.60\pm 0.08\%$. Black points violate the specification. Red and blue points are the points sampled in nesting $k-1$ that are outside or inside of $\mathcal{L}_k$, respectively.}
     \label{fig:perching_ver_ic}
\end{figure}

we allow stochasticity in the actual initial location $(x,z)'$ of the plane w.r.t the conditions used for finding the optimal trajectory and the LQR-tree controllers. Meaning, the trajectory and controller are computed once for the nominal conditions, and remain fixed. 
\begin{flalign}
\label{eqn:perching_noises_dist}
    {\bm w}_0=[x_0, z_0]'\sim\mathcal{N}\left([-3.5, 0.1]', \operatorname{diag}([0.001^2,0.02^2])\right),&&
\end{flalign}
and the specification:
\begin{align}
\label{eqn:perching_specification}
    \varphi_p := \Diamond_{[0,66]}\left(\lVert \text{Plane Pose}-\text{Wire Pose}\rVert _1\leq \varepsilon\right),
\end{align}
where $\varepsilon=0.05$, $T=0.66$sec and $\Delta t=0.01$sec. 
Fig.~\ref{fig:perching_ver_ic} shows the comparison of STL-based ESS and MC where black dots represent violating $\varphi_p$. 
The probability to fail with STL-based ESS is $p_f^{\varphi_p}=0.85\%$ and with MC $p_f^{\varphi_p}=0.60\pm 0.08\%$ with $8,880$ simulations.

\subsubsection{Actuation noises} 
\begin{figure}
     \centering
     \begin{subfigure}[b]{0.49\columnwidth}
         \centering
         \includegraphics[width=\columnwidth]{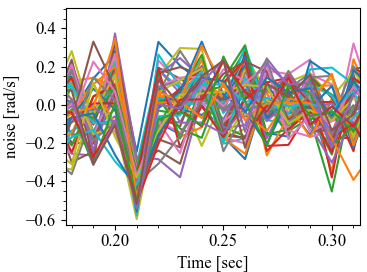}
         \caption{}
         \label{fig:perching_act_noise_analysis}
     \end{subfigure}
     \hfill
     \begin{subfigure}[b]{0.49\columnwidth}
         \centering
         \includegraphics[width=\columnwidth]{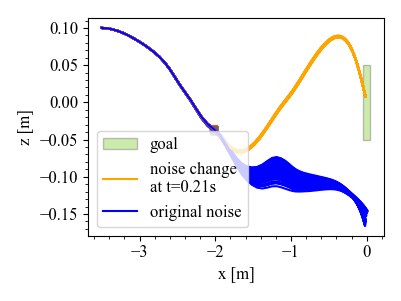}
         \caption{}
         \label{fig:perching_diff_noise}
     \end{subfigure}
        \caption{(a) Actuation noise for $\varphi_p$-violating trajectories. A common feature through all of the signals is a negative noise at $t=0.21$sec. (b) Trajectories with $\varphi_p$-violating noise in blue and trajectories with a single change in the noise in orange, where we flip the sign of the noise at $t=0.21$sec.}
     \label{fig:perching_act_noise}
\end{figure}

the plane has only one actuator, which is controlling the elevation rate of the hinged aileron. Noise ${\bm w}\in\mathbb{R}^{67}$ where $w_t\sim\mathcal{N}\left(0,0.15^2\right)$ is added to the actuator pitch rate at every time step. The probability to fail with STL-based ESS is $p_f^{\varphi_p}=0.026\%$ with $18,189$ simulations and with MC $p_f^{\varphi_p}=0.13\pm 0.02\%$ with $50k$ simulations. Fig.~\ref{fig:perching_act_noise_analysis} shows one possible diagnosis of failure reasons which is a drop to a negative noise rate at the critical time of $t=0.21$sec. 

\section{Discussion}
\label{sec:discussion}

\noindent In this work we provide a probabilistic method to verify robotic systems w.r.t safety and liveness properties in the form of an STL specification. The systems have internal noise and disturbances or perhaps acted upon externally by the environment, and we want to find the probability that the system would fail to comply with the specification. In this work, we provide bounds on the error that we may experience due to the probabilistic nature of the method. We also provide another algorithm to efficiently sample from the active domain when the system is non-linear or black-box. Finally, we utilize this method to efficiently find the gradients of the probability w.r.t any parameter of the system, a feature that enables us to find the optimal parameter that would minimize the probability of failure. 

The use of STL specifications allows for the validation of a much richer, more complex, and more realistic set of requirements and properties as well as temporal relations. As mentioned in the introduction, other works usually provide some cost or mathematical function, such as the minimal distance to obstacles, to try to find a set of noises that would violate them. Obtaining the probability gives us a notion of understanding the likelihood of failing, as opposed to finding a failing example or the most likely failing example. 

Note that we use our method on a computer in simulation, where the simulation is usually integrated and in discrete time. One has to expect the possible introduction of errors from the continuous system due to the numerical integration of the states and the discrete-time nature of their representation.

While the method works well for linear systems because we can get the ellipse intersections analytically, the method may not work well in all black-box systems. For example, systems that are chaotic (a small change in noise parameters leads to completely different outputs) or completely random or specifications that try to capture events that are too abrupt and sparse. In these cases though, it is extremely hard to estimate the probability even with MC or any other methods.

Finally, this method may be used with malicious intentions, such as to find an autonomous car's weak spots and exploit them. However, we mention that first, the car's simulation and noise distribution should not be accessible to the attacker. And second, this method is not the optimal way to obtain this information, and finding possible exploitations is much faster with the related works discussed in this paper.

For future work, we intend to (1) examine the perception errors of cameras and their effects on the probability to fail a task. (2) provide bounds on the error that the non-linear formulation may introduce when sampling with ESS. (3) provide an approximation method to synthesize new controls for non-linear systems.

\addtolength{\textheight}{-1.5cm}   


\bibliographystyle{IEEEtran}
\bibliography{IEEEabrv,refs.bib}

\end{document}